%% file: eccv_labelgan_for_review.tex
\documentclass[article]{llncs}
\usepackage{graphicx}
\usepackage{comment}
\usepackage{amsmath,amssymb} 
\usepackage{color}

\usepackage[width=122mm,left=12mm,paperwidth=146mm,height=193mm,top=12mm,paperheight=217mm]{geometry}
\usepackage{times}
\usepackage{epsfig}
\usepackage{graphicx}
\usepackage{amsmath}
\usepackage{amssymb}
\usepackage{dsfont}
\graphicspath{{figures/}}
\usepackage{dsfont}
\usepackage[dvipsnames]{xcolor}
\usepackage{cite}
\usepackage{multirow}
\usepackage{adjustbox}

\usepackage{color, colortbl}
\definecolor{Gray}{gray}{0.9}
\newcolumntype{g}{>{\columncolor{Gray}}c}

\DeclareRobustCommand\onedot{\futurelet\@let@token\@onedot}
\def\@onedot{\ifx\@let@token.\else.\null\fi\xspace}

\usepackage[pagebackref=true,breaklinks=true,colorlinks,bookmarks=false]{hyperref}

\begin{document}
\pagestyle{headings}
\mainmatter
\def\ECCVSubNumber{5756}  

\title{Semantics-Aware Image to Image Translation and Domain Transfer
} 

\titlerunning{SemGAN}
%
\author{Pravakar Roy\inst{1} \and
Nicolai H{\"a}ni\inst{1} \and
Jun-Jee Chao\inst{1}\and
Volkan Isler\inst{1}}
\authorrunning{Roy et al.}
%
\institute{University of Minnesota, Minneapolis, MN 55455, USA 
\email{\{royxx268,haeni001,chao0107,isler\}@umn.edu}}

\maketitle



\input{abstract}

\section{Introduction}\label{sec:intro}
\input{intro}
\section{Related Work}\label{sec:relwork}
\input{relwork}
\section{Semantics-Aware GAN}\label{sec:formulation}
\input{formulation}
\section{Implementation}\label{sec:implementation}
\input{implementation}

\section{Experimental Evaluation}\label{sec:rexperiments}
\input{results}
\section{Conclusion}\label{sec:conc}
\input{conc}
\section{Acknowledgement}
This work is supported in part by NSF grant \# 1722310 and USDA
NIFA MIN-98-G02.
\clearpage
{\small
\bibliographystyle{splncs04}

\input{output}
}
\input{eccv_supplementary}
\end{document}

%% file: abstract.tex
\begin{abstract}
Image to image translation is the problem of transferring an image from a source domain to a different (but related) target domain. We present a new unsupervised image to image translation technique that leverages the underlying semantic information for object transfiguration and domain transfer tasks. Specifically, we present a generative adversarial learning approach that jointly translates images and labels from a source domain to a target domain. Our main technical contribution is an encoder-decoder based network architecture that jointly encodes the image and its underlying semantics and translates both individually to the target domain. Additionally, we propose \textit{object transfiguration} and \textit{cross domain semantic consistency} losses that preserve semantic labels. Through extensive experimental evaluation, we demonstrate the effectiveness of our approach as compared to the state-of-the-art methods on unsupervised image-to-image translation, domain adaptation, and object transfiguration. 

\keywords{Generative Adversarial Network, Unsupervised Image to Image Translation, Domain Adaptation}
\end{abstract}

%% file: intro.tex
Semantic segmentation, i.e., assigning a per-pixel class label, is a way of understanding regions and structure of an image. It is a common preprocessing step in robotics applications~\cite{wolf_enhancing_2016,shvets_automatic_2018}, autonomous driving~\cite{chen_monocular_2016,teichmann_multinet_2018}, and medical image processing~\cite{xue_segan_2018,stoyanov_unet_2018}. Similar to many other visual perception tasks, semantic segmentation has seen fast progress with the adaptation of deep convolutional neural networks (CNN's). For dense prediction tasks, including semantic segmentation,  collecting large-scale and diverse datasets is a challenge. For example, it is non-trivial to acquire real-world examples of driving scenes with varying weather conditions because obtaining pixel-level annotation for all variations is very hard. Synthetic data promises to be a viable alternative, where the environment is modifiable, and labels are obtained for free. However, using synthetic data to train networks that perform well on real data is challenging due to the domain gap between the two.  Thus, developing algorithms that can transfer knowledge across synthetic and real data has become popular. Nevertheless, due to the domain gap problem, these algorithms often fail to generalize to new datasets.

\begin{figure}[!hbpt]
    \centering
    \def\svgwidth{\columnwidth}
    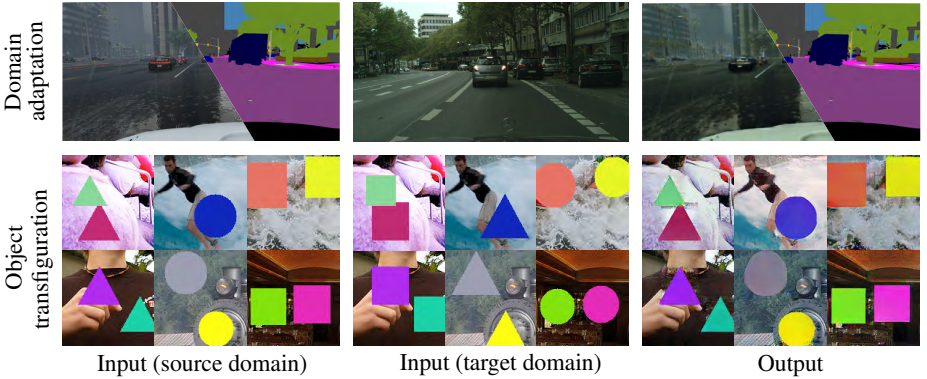
    \caption{\textbf{Applications of the proposed method}. Our method has applications in domain adaptation (top row) and geometric transfiguration of individual objects (bottom row).}
    \label{fig:intro}
\end{figure}

The task of finding a mapping to translate images from a source domain to a target domain is known as \emph{image to image translation}. Existing methods tackle this problem by either pairing the images from the two domains if such pairs are available~\cite{eigen_predicting_2015, isola2017image, wang_video--video_2018, xian_texturegan:_2018} or by translating samples on the distribution level~\cite{zhu_unpaired_2017, yi_dualgan:_2017, tomei_art2real:_2018}. However, these image-to-image translation techniques often fail when the mapping function includes significant geometric changes as there is no explicit constraint that considers the underlying representation of the scene. While several methods have been proposed to alleviate this issue via self attention~\cite{alami_mejjati_unsupervised_2018} or through enforcing label consistency~\cite{li_semantic-aware_2018}, the problem remains challenging since these methods may only depend on soft constraints.

\textbf{Our contribution.}  In this paper, we present SemGAN, an unsupervised image to image translation algorithm that preserves class labels during the translation process. Our solution consists of two main modules: 1) an encoder network that jointly estimates a latent variable $z$ from the combined image/class input and 2) two separate decoder networks that reconstruct the image and class outputs. As shown in Figure~\ref{fig:overview}, our core idea is to translate input images and class maps jointly. We compare the translated images with the target distribution to adapt the image style, and introduce \textit{cross-domain consistency} and \textit{object transfiguration} losses to preserve the semantic labels. Our approach applies to a variety of object transfiguration and unsupervised domain adaptation tasks (Figure~\ref{fig:intro}). 

We perform extensive qualitative and quantitative evaluation of our approach for domain adaption, unsupervised image to image translation, and object transfiguration tasks on multiple datasets. Results indicate that our approach achieves state-of-the-art performance against existing methods. Our datasets, model weights, and code will be made available upon acceptance of this paper.


%% file: intro_figure.pdf_tex
\begingroup%
  \makeatletter%
  \providecommand\color[2][]{%
    \errmessage{(Inkscape) Color is used for the text in Inkscape, but the package 'color.sty' is not loaded}%
    \renewcommand\color[2][]{}%
  }%
  \providecommand\transparent[1]{%
    \errmessage{(Inkscape) Transparency is used (non-zero) for the text in Inkscape, but the package 'transparent.sty' is not loaded}%
    \renewcommand\transparent[1]{}%
  }%
  \providecommand\rotatebox[2]{#2}%
  \newcommand*\fsize{\dimexpr\f@size pt\relax}%
  \newcommand*\lineheight[1]{\fontsize{\fsize}{#1\fsize}\selectfont}%
  \ifx\svgwidth\undefined%
    \setlength{\unitlength}{973.44178688bp}%
    \ifx\svgscale\undefined%
      \relax%
    \else%
      \setlength{\unitlength}{\unitlength * \real{\svgscale}}%
    \fi%
  \else%
    \setlength{\unitlength}{\svgwidth}%
  \fi%
  \global\let\svgwidth\undefined%
  \global\let\svgscale\undefined%
  \makeatother%
  \begin{picture}(1,0.40058193)%
    \lineheight{1}%
    \setlength\tabcolsep{0pt}%
    \put(0.09791277,0.00384632){\color[rgb]{0,0,0}\makebox(0,0)[lt]{\lineheight{1.25}\smash{\begin{tabular}[t]{l}Input (source domain)\end{tabular}}}}%
    \put(0.42949887,0.00431794){\color[rgb]{0,0,0}\makebox(0,0)[lt]{\lineheight{1.25}\smash{\begin{tabular}[t]{l}Input (target domain)\end{tabular}}}}%
    \put(0.82194372,0.00419184){\color[rgb]{0,0,0}\makebox(0,0)[lt]{\lineheight{1.25}\smash{\begin{tabular}[t]{l}Output\end{tabular}}}}%
    \put(0.01510448,0.09116784){\color[rgb]{0,0,0}\rotatebox{89.44905}{\makebox(0,0)[lt]{\lineheight{1.25}\smash{\begin{tabular}[t]{l}Object \end{tabular}}}}}%
    \put(0.01381761,0.28391177){\color[rgb]{0,0,0}\rotatebox{89.44905}{\makebox(0,0)[lt]{\lineheight{1.25}\smash{\begin{tabular}[t]{l}Domain \\\end{tabular}}}}}%
    \put(0,0){\includegraphics[width=\unitlength,page=1]{intro_figure.pdf}}%
    \put(0.0404113,0.27230205){\color[rgb]{0,0,0}\rotatebox{91.37005}{\makebox(0,0)[lt]{\lineheight{1.25}\smash{\begin{tabular}[t]{l}adaptation\end{tabular}}}}}%
    \put(0.04510987,0.04844536){\color[rgb]{0,0,0}\rotatebox{90.91854}{\makebox(0,0)[lt]{\lineheight{1.25}\smash{\begin{tabular}[t]{l}transfiguration\end{tabular}}}}}%
  \end{picture}%
\endgroup%

%% file: relwork.tex
\textbf{Unsupervised image-to-image translation.} Image-to-image translation is the process of learning a generative model to translate images from a given source distribution to a different (but related) target distribution. In the semantic segmentation case, we would like to leverage synthetic samples (for which segmentation is free) to classify real images.

To close the domain gap between the two distributions, Isola et al.~\cite{isola2017image} proposed ``pix2pix'', a conditional Generative Adversarial Network (GAN)~\cite{goodfellow_generative_2014} that uses paired samples. Extensions of this approach have been used in a variety of applications~\cite{eigen_predicting_2015,wang_video--video_2018,xian_texturegan:_2018}. However, acquiring paired data is difficult for most scenarios. Unsupervised image-to-image translation, powered by identity ~\cite{taigman_unsupervised_2016} and cyclic consistency losses~\cite{zhu_unpaired_2017} have enabled domain adaptation for many interesting applications~\cite{yi_dualgan:_2017,gonzalez-garcia_image--image_2018,benaim2017one,zhang_harmonic_2019,amodio_travelgan:_2019,murez_image_2017, hong_conditional_2018, zhang_collaborative_2018, shen_wasserstein_2017, hassan_unsupervised_2018, long_conditional_2018, hosseini-asl_augmented_2019}. Most of these works include adversarial learning frameworks together with a combination of identity losses and cyclic consistency to avoid optimization instabilities during training. While these approaches can translate from one image domain to another, the underlying semantics are lost if the network is allowed to manipulate an object's geometry. In this work, we show that we can perform domain adaptation while preserving these semantic label maps.

\textbf{Semantically consistent domain adaptation.}
Unsupervised image-to-image translation methods often change images in such a way that they become inconsistent with their labels. There exist two main approaches to address this challenge: 1) task preserving methods, and 2) semantics preserving methods. 
Task preserving methods aim to learn a task network on the source domain that also performs well on the target domain, often by aligning the feature distributions between the two domains. For semantic segmentation, existing methods proposed to align the feature distributions ~\cite{hoffman_fcns_2016,huang_domain_2018,tsai_learning_2018} or to employ cross-domain consistency by penalizing inconsistent predictions of the task network~\cite{chen_crdoco_2019}.

The second line of work is unsupervised image-to-image translation that aims at translating pixel appearance while preserving the underlying semantics. Li et al.~\cite{li_semantic-aware_2018} proposed a network with a Sobel filter loss that is convolved with the semantic label in the source domain and the translated image to preserve the boundaries between the classes. Other approaches~\cite{hoffman_cycada:_2017,ramirez_exploiting_2018} proposed a task network for semantic segmentation. The task network segments the translated images, and a task prediction loss penalizes any changes made to the semantic labels. Cherian and Sullivan~\cite{cherian_sem-gan:_2018} proposed to learn two task networks, one for the source and one for the target domain. During training, the network learns by minimizing disagreement between the two task networks.
Introducing a task network in addition to the generative image-to-image translation model is computationally expensive. Additionally, these networks restrict the geometric changes that a network is allowed to perform rendering them useless for translation requiring substantial geometric changes, as in Sheep $\to$ Giraffe. Our model does not rely on any additional task network. Instead, we preserve the semantics for domain transfer tasks with our cross-domain semantic consistency loss term. 

\textbf{Object transfiguration and semantic manipulation}
Most of the image to image translation methods discussed so far are capable of modifying low-level content of images, such as transferring colors or textures. They fail to perform substantial semantic changes between objects (e.g., transform cats into dogs). This failure is due to the underlying assumption that the scenes and the contained objects are similar in geometric composition across both domains. Liang et al.~\cite{liang_generative_2017} proposed to address this issue by optimizing a conditional generator and several semantic-aware discriminators. Another idea~\cite{alami_mejjati_unsupervised_2018} relies on modeling an attention map to extract the foreground objects. 
Mo et al. used instance-level information~\cite{mo_instance-aware_2019,shen2019towards} to translate object instances from the source to the target domains. Instead of using instance mask, Tang et al.~\cite{tang2019cycle} used keypoints to generate novel human pose images.  

In this paper, we show that we can obtain better performance on the object transfiguration task with a change to how images and classes are encoded. We present an encoder-decoder based generator architecture that can translate both the input image and corresponding semantic map to the target domain jointly in a single forward pass of the network.\\

%% file: formulation.tex

We consider the tasks of domain adaptation and object transfiguration with semantic consistency. In this setting, we assume access to a large collection of images and labels from a source domain $X_S$, and a small set of images and labels from a target domain $X_T$. Our goal is to learn inverse mappings $G_{S\to T}:S \to T$ and $G_{T\to S}: T \to S$ that can reliably fool a pair of adversarial discriminators $D_S, D_T$, while keeping the image consistent with its associated label map. 

\begin{figure}[!hbpt]
     \centering
     \includegraphics[width=\textwidth]{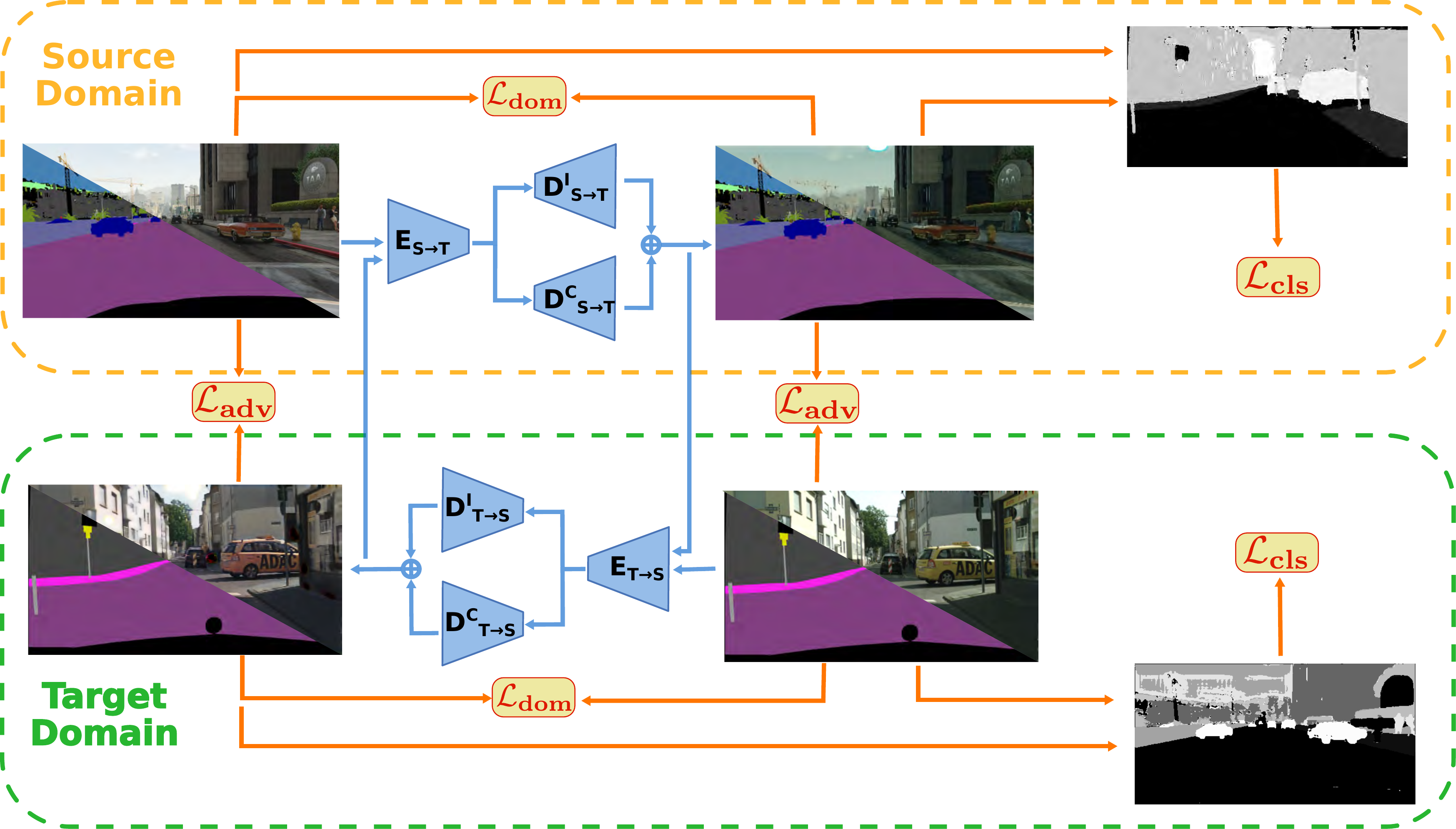}
     \caption{\textbf{Overview of the proposed method.} Our model is composed of two main networks: a joint image/class generator to translate from source to target domain and a network to translate from target to source domain. The networks learn to jointly encode the image/class map pair and decode them into a translated pair that looks like the second domain. Our main contribution lies in the use of a joint encoder, split decoder architecture and the cross-domain consistency loss $\mathcal{L}_{dom}$ as well as the object transfiguration loss $\mathcal{L}_{cls}$. Note that the cross-domain consistency loss $\mathcal{L}_{dom}$ is only used for domain adaptation and object transfiguration loss $\mathcal{L}_{cls}$ is only used for object transfiguration. We do not show all loss terms for clarity.}
     \label{fig:overview}
\end{figure}

To achieve this goal, we present an end-to-end trainable network which is composed of two main components: 1) a joint image/class map encoder and 2) two domain-specific decoders, one each for the image and the class map. 
As shown in Figure~\ref{fig:overview}, the proposed network takes and image $I_S$ and corresponding class map $C_S$ from the source domain and an image $I_T$ together with the corresponding class map $C_T$ from the target domain as inputs. The image and class map are concatenated into a tensor $R$ of size $R^{(3+M)\times H \times W}$, where $(H,W)$ are the image size and $M$ is the number of classes. We first use the encoder network to generate joint latent representations $z_S = E_{S\to T}(R_S)$ and $z_T = E_{T\to S}(R_T)$. The decoder networks receive this latent representation $z_S, z_T$ as input and produce the adapted image and class maps independently as $I_{S\to T} = D_{S \to T}^I(z_S)$, $C_{S\to T} = D_{S \to T}^C(z_S)$ (in the source domain) and $I_{T\to S} = D_{T \to S}^I(z_T)$, $C_{T\to S} = D_{T \to S}^C(z_T)$ (in the target domain). 

\subsection{Objective Function}
The complete objective function $\mathcal{L}$ for the training of our proposed network consists of five loss terms. First, the joint image/class map adversarial loss $\mathcal{L}_{adv}$ aligns the image and class distributions between the translated image/class pairs and the image/class pairs in the corresponding domain. Second, the identity loss $\mathcal{L}_{idt}$ regularizes the generator networks $G_{S\to T}$ and $G_{T\to S}$ to perform the identity function when being shown an image/label map pair from the adapting domain (i.e. $I_T, C_T \approx G_{S\to T}(I_T, C_T))$. Third, the cyclic consistency loss, or reconstruction loss $\mathcal{L}_{rec}$, regularizes the generator networks to perform self-reconstruction when translating an image from one domain to another, followed by a reverse translation. Fourth, the class preserving loss $\mathcal{L}_{cls}$ preserves corresponding semantic labels between the original and the adapted image. Fifth, the cross-domain loss $\mathcal{L}_{dom}$ regularizes the generator networks the keep the semantic label maps as of the adapted image as close to the source images class map as possible. The complete training objective $\mathcal{L}$ is defined as
\begin{equation}
    \mathcal{L} = \lambda_{adv} \cdot \mathcal{L}_{adv} + \lambda_{idt} \cdot \mathcal{L}_{idt} + \lambda_{rec} \cdot \mathcal{L}_{rec} + \lambda_{cls} \cdot \mathcal{L}_{cls} + \lambda_{dom} \cdot \mathcal{L}_{dom}
\end{equation}

where $\lambda_{adv}, \lambda_{idt}, \lambda_{rec}, \lambda_{cls}$ and $\lambda_{dom}$ are the hyper-parameters controlling the relative importance of the different losses. Below, we discuss each term of the loss functions in detail. In Section~\ref{sec:rexperiments}, we present qualitative and quantitative ablations of the individual loss terms in our objective function. 

\subsubsection{Object transfiguration loss.}\label{subsec:cls_loss}
To preserve the semantic elements of an image (class maps) we introduce a \textit{object transfiguration loss} $\mathcal{L}_{cls}$. We first translate a image/class map pair $R$ to the other domain, namely $R_{S\to T} = G_{S\to T}(R_S)$ for the source domain and $R_{T\to S} = G_{T\to S}(R_T)$ for the target domain. We create a pixel-indicator map $P$ that captures pixels that changed their semantic values during translation $P_{S\to T} = \overline {C_S \cup C_{S\to T}}$ for the translated source class map and $P_{T\to S} = \overline {C_T \cup C_{T\to S}}$ for the translated target class map. We use this pixel-indicator map to regularize our generator network by penalizing pixels where the pixel indicator function is 1. Specifically, we define our object transfiguration loss $\mathcal{L}_{cls}$ as

\begin{equation}
\begin{aligned}
\mathcal{L}_{cls}(X_S, X_T;  G_{S\to T}, G_{T\to S}) = \mathbb{E}_{R_S \sim X_S}[P_{S\to T}\odot||I_S - I_{S\to T}||_1]\\
+ \mathbb{E}_{R_T \sim X_T}[P_{T\to S}\odot||I_T - I_{T\to S}||_1]
\end{aligned}
\label{eq:lossclasspreseve}
\end{equation}

where $\odot$ is an element-wise multiplication.

\subsubsection{Cross-domain semantic consistency loss.}
For domain translation applications, it is desired that the underlying geometry of the environment is preserved in the translation process. To encourage preserving geometry in such cases, we add a \textit{cross domain semantics preservation loss} $\mathcal{L}_{dom}$ term to our objective. We regularize semantic labels using the standard cross-entropy loss for multi-class classification. More precisely, we define the cross-domain semantic consistency loss $\mathcal{L}_{dom}$ as

\begin{equation}
\begin{aligned}
\mathcal{L}_{dom}(X_S, X_T;  G_{S\to T}, G_{T\to S}) = - \sum_{c=1}^M y_{o,c \in C_S} G_{S\to T}(C_S)) \\
- \sum_{c=1}^M y_{o,c \in C_T} \log(G_{T\to S}(C_T))
\end{aligned}
\label{eq:losssem}
\end{equation}

where, $M$ is the total number of classes, $y_{o, c}$ is a binary indicator (0 or 1) if the class label $c$ is the correct classification for observation $o$ and the predicted class probabilities $G_{S\to T}(C_S), G_{T\to S}(C_T)$.

\subsubsection{Joint image/class-level adversarial loss.}\label{subsec:advloss}
The standard adversarial loss for GAN networks~\cite{goodfellow_generative_2014} aims to align two image distributions between image $I_S$ and the translated one $I_{S\to T}$. We modify this loss to jointly align the distributions between image/class map pairs $R_S$ and the translated pairs $R_{T\to S}$ using two joint image/class-level discriminators $D_S$ (for the source domain) and $D_T$ (for the target domain). We generate translated image/class map pairs $R_{S\to T} = I_{S\to T}, C_{S\to T}$ in the source domain and $R_{T\to S} = I_{T\to S}, C_{T\to S}$ in the target domain. Then, we define the joint image/class map adversarial loss as

\begin{equation}
\begin{aligned}
\mathcal{L}_{adv}(X_S, X_T; G_{S\to T}, D_{S}) = \mathbb{E}_{R_S \sim X_S}[\log(D_S(R_S))] \\
+ \mathbb{E}_{R_T \sim X_T} [\log(1-D_S(R_{T\to S})]
\end{aligned}
\label{eq:lossadv}
\end{equation}

Similarly, we have another image/class-level adversarial loss in the target domain as $\mathcal{L}_{adv}(X_T, X_S; G_{T\to S}, D_T)$.

\subsubsection{Identity loss.}\label{subsec:idtloss}
We apply an identity loss $\mathcal{L}_{idt}$ that regularizes the generator networks $G_{S\to T}$ and $G_{T\to S}$ to perform the identity function when being shown an image from the target domain . Namely, we want that an image/class map pair $R = (I, C)$ is roughly equal to itself if shown to generator network such that $R_S \approx G_{T\to S}(R_S)$ (in the source domain) and $R_T \approx G_{S\to T}(R_T)$ (in the target domain). We use the $L_1$ norm for the identity loss, as proposed by Taigman et al.~\cite{taigman_unsupervised_2016}. To account for the semantic labels we use the standard cross-entropy loss for multi-class classification. More precisely, we define the identity loss $\mathcal{L}_{idt}$ as

\begin{equation}
\begin{aligned}
\mathcal{L}_{idt}(X_S,X_T; G_{S\to T}, G_{T\to S}) = E_{I_S \sim X_S}[||G_{T\to S}(I_S) - I_S||_1]\\
+ E_{I_T \sim X_T}[||G_{S\to T}(I_T) - I_T||_1]\\
- \sum_{c=1}^M y_{o,c\in C_S} \log(G_{T\to S}(C_S)) \\
- \sum_{c=1}^M y_{o,c \in C_T} \log(G_{S\to T}(C_T))
\end{aligned}
\end{equation}

where, $M$ is the total number of classes, $y_{o, c}$ is a binary indicator (0 or 1) if the class label $c$ is the correct classification for observation $o$ and the predicted class probabilities $G_{T\to S}((C_S)), G_{S\to T}((C_T))$.

\subsubsection{Reconstruction loss.}\label{subsec:cycleloss}
We use an image/class reconstruction loss $\mathcal{L}_{rec}$ to regularize the training of our generator network. Similar to previous work, we use the cyclic consistency property~\cite{zhu_unpaired_2017}, which states that when we translate inputs from one domain to another, followed by the reverse translation, we should obtain the original input. Namely, $G_{T\to S}\left(G_{S\to T}(R_S)\right) \approx R_S$ for any image/class pair $R_S$ in the source domain and $G_{S\to T}\left(G_{T\to S}(R_T)\right) \approx R_T$ for any $R_T$ in the target domain. We enforce image consistency across the two mappings $G_{S\to T}, G_{T\to S}$ using the $L_1$ norm between the original and the reconstructed image. To account for the semantic labels, we use the standard cross-entropy loss for multi-class classification. More precisely, we define the reconstruction loss $\mathcal{L}_{rec}$ as

\begin{equation}
\begin{aligned}
\mathcal{L}(X_S, X_T; G_{S\to T}, G_{T\to S}) =  \mathbb{E}_{I_S \sim X_S}[||G_{T\to S}(G_{S\to T}(I_S)) - I_S||_1] \\
+ \mathbb{E}_{I_T \sim X_S}[||G_{S\to T}(G_{T\to S}(I_T)) - I_T||_1] \\
- \sum_{c=1}^M y_{o,c\in C_S} \log(G_{T\to S}(G_{S\to T}(C_S))) \\
- \sum_{c=1}^M y_{o,c \in C_T} \log(G_{S\to T}(G_{T\to S}(C_T)))
\end{aligned}
\label{eq:lossrec}
\end{equation}

where, $M$ is the total number of classes, $y_{o, c}$ is a binary indicator (0 or 1) if the class label $c$ is the correct classification for observation $o$ and the predicted class probabilities $G_{T\to S}(G_{S\to T}(C_S)), G_{S\to T}(G_{T\to S}(C_T))$.

Based on the aforementioned objective function, we aim to solve for translation networks $G_{S\to T}, G_{T\to S}$ by optimizing the following min-max problem:

\begin{equation}
G_{S\to T}^*, G^*_{T\to S} = \min_{\substack{G_{S\to T} \\ G_{T\to S}}} \max_{\substack{D_S^{R}\\ D_{T}^R}}\mathcal{L}
\end {equation}

To be specific, the identity loss $\mathcal{L}_{idt}$ and the reconstruction loss $\mathcal{L}_{rec}$ function as regularizers for our objective. The joint image/class adversarial loss $\mathcal{L}_{adv}$ drives the image to image translation from the source domain to the target domain. The proposed object transfiguration $\mathcal{L}_{cls}$ and cross-domain semantic consistency loss $\mathcal{L}_{dom}$, in contrast, keeps the background consistent and aligns the semantic label maps in across different domains.

%% file: implementation.tex
We implement our network on top of the PyTorch version of Cycle-GAN~\cite{zhu_unpaired_2017}. In particular, we used building blocks from the  ResNet 9-blocks generator for our generator networks $G_{S\to T}$ and $G_{T\to S}$. In addition to our new method, we replace the used deconvolution layers with upsampling layers followed by regular convolutions for the decoders to eliminate checkerboard artifacts. The network receives concatenated images and semantic maps as inputs. Before concatenation, we convert the semantic maps to one-hot encoding, normalize, and zero-center the inputs. As a result, the dimension of the network input depends on the total number of classes. For an image with dimension, $h\times w \times c$ and class map containing $M$ classes, the dimension of the input is $h \times w \times c \times M$. For the activation layers of the two decoders (image and semantic labels), we used $\tanh$ and soft-max non-linearities, respectively.For our discriminators $D_{S}$ and $D_{T}$, we used the PatchGAN~\cite{isola2017image} network. For a detailed description of the network and used parameters for each experiment, please consult the supplementary document.

%% file: results.tex
\subsection{Unsupervised Image-to-Image Translation}
We present experimental results for unsupervised image-to-image translation for two different settings: 1) \textit{synthetic-to-real} domain adaptation of street scenes from the GTA~\cite{richter_playing_2016} to the Cityscapes~\cite{cordts_cityscapes_2016} datasets, and 2) \textit{real-to-real} image-to-image translation of images extracted from the COCO~\cite{lin2014microsoft} dataset.

\subsubsection{Synthetic-to-real adaptation}
\flushleft{\textbf{Dataset.}}
The \textit{synthetic} GTA5~\cite{richter_playing_2016} dataset contains $24,964$ images with pixel-level annotations of $19$ categories. The Cityscapes~\cite{cordts_cityscapes_2016} dataset contains $2,975$ images of European cities, annotated with $34$ categories. Following Hoffman et al.~\cite{hoffman_cycada:_2017}, we use the GTA5 dataset and adapt it to the Cityscapes training set. For training our adaptation network, we use $500$ images with class labels from both sides. In an effort to minimize the training time, we restrict the adapted GTA5 dataset to only $5,000$ images, sampled uniformly.

\textbf{Evaluation Protocol.}
We use the adapted GTA5 images to train a task network that we test on the Cityscapes validation set with 500 images. For evaluation, we use class intersection-over-union (IoU), mean intersection-over-union (mIoU), and pixel accuracy as evaluation metrics. 

\textbf{Task Network.} We evaluate our proposed method using a dilated residual network-26 (DRN-26)~\cite{long2015learning}. The network is trained on $5,000$ translated images selected at random and tested on the Cityscapes test dataset.

\begin{figure}[!hbpt]
    \centering
    \def\svgwidth{\columnwidth}
    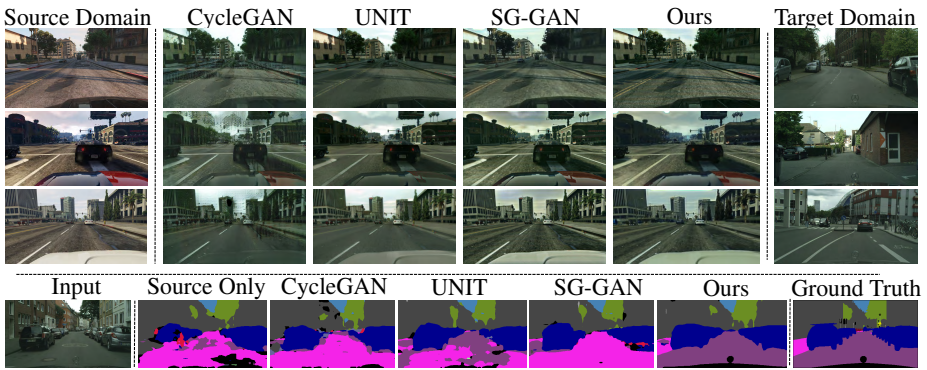
    \caption{\textbf{Visual results for domain adaptation and semantic segmentation}. We show sample results for the \textit{synthetic-to-real} domain adaptation case from the GTA5 to the Cityscapes dataset (top three rows). We use the translated images to train a task network and test it on the Cityscapes test set (bottom row).}
    \label{fig:gta2city_qual}
\end{figure}

\textbf{Results.} We compare our approach with the state-of-the-art baselines~\cite{zhu_unpaired_2017,liu2017unsupervised,li_semantic-aware_2018,mo_instance-aware_2019} and ablations of our objective function. Additionally, we compare our approach to the source only case (no adaptation) and the oracle case (target labels are available). Table~\ref{tab:gta} presents the quantitative results. The results show that our method performs favorably in both mIoU score and pixel accuracy, outperforming the next best method by 5.4\% in mean IoU and 13.5\% in pixel accuracy. We show that the proposed cross-domain semantic consistency loss $\mathcal{L}_{dom}$ is critical for the improved performance (e.g., adding $\mathcal{L}_{dom}$ improves the mean IoU by 8\% and pixel accuracy by 23\%). Figure~\ref{fig:gta2city_qual} shows qualitative examples of the proposed domain translation together with a comparison of the resulting semantic segmentation quality of the task network.

\begin{table}[htbp!]
\caption{\textbf{Experimental results of synthetic-to-real adaptation for semantic segmentation}. We present IoU and pixel accuracy (higher numbers are better) and indicate the top results as \textbf{bold}.} \label{tab:gta}
\begin{adjustbox}{width=1\textwidth}
\centering
\begin{tabular}{l|cccccccccccccccccccc|c|c}
\hline
\multicolumn{23}{c}{\textbf{GTA $\to$ Cityscapes}}\\
\hline
Method & \rotatebox{90}{Road} & \rotatebox{90}{Sidewalk} & \rotatebox{90}{Building} & \rotatebox{90}{Wall} & \rotatebox{90}{Fence} & \rotatebox{90}{Pole} & \rotatebox{90}{Traffic Light} & \rotatebox{90}{Traffic Sign} & \rotatebox{90}{Vegetation} & \rotatebox{90}{Terrain} & \rotatebox{90}{Sky} & \rotatebox{90}{Person} & \rotatebox{90}{Rider} & \rotatebox{90}{Car} & \rotatebox{90}{Truck} & \rotatebox{90}{Bus} & \rotatebox{90}{Train} & \rotatebox{90}{Motorbike} & \rotatebox{90}{Bicycle} & \rotatebox{90}{Unlabeled} & \rotatebox{90}{mean IoU} & \rotatebox{90}{Pixel acc.} \\
\hline
Source only & 20.85 & 9.17 & 51.28 & 1.79 & 0.26 & 4.73 & 0.67 & 0.00 &    57.14 &    5.20 & 48.90 & 3.99 & 0.00 & 19.14 &    0.38 &    0.01 &    0.00 &    0.00 &    0.00 &    9.39 & 11.65 & 46.34 \\
CycleGAN~\cite{zhu_unpaired_2017} & 59.30 &    16.10 &    53.25 &    2.48 &     0.26 &    5.96 &    0.79 &    0.03 &    54.69 &    4.53&    46.87 &    3.11 &    0.00 &    36.56 &    0.91 &    0.16 &    0.00 &    0.01 &    0.01 &    29.66 & 15.73 & 65.27\\
SG-GAN~\cite{li_semantic-aware_2018} & 35.01 & 10.92 & 56.84 & \textbf{2.71} & 0.58 & 5.74 &    0.92 & 0.01 & 60.38 & 5.61 &    53.56 &    3.30 &    0.00 &    34.59 &    0.90 &    0.10 &    0.00 &    0.01 &    0.00 &    22.77 & 14.07 & 55.68\\
UNIT~\cite{liu2017unsupervised} & 55.99 & 17.84 & 55.51 &    2.51 &    0.81 &    6.05 &    1.01 &    0.05 &    61.33 &    7.21 &    49.80 &    4.82 &    0.00 &    36.97 &    \textbf{1.21} &    0.26 &    0.00 &    0.20 &    0.16 &    25.52 & 16.36 & 64.86\\
Ours w/o rec. loss &  30.84 & 20.10    & 10.41 & 0.47 & 0.04 & 0.96 & 0.13    & 0.28    & 4.16 & 1.63 & 24.80 & 0.08 & 0.00 & 1.73 & 0.02 & 0.13 & 0.06    & 0.01 & 0.00 & 4.90 & 4.13 & 26.47\\
Ours w/o idt. loss &  42.25 & 3.21 & 31.21 & 0.94 &    0.32 & 2.73 & 0.00 & 0.16 & 25.40 & 2.48 & 13.89 & 2.79 & 0.00 & 4.46 & 0.10 & 0.41 & 0.00 & 0.00 & 0.00 &    8.06 & 6.92 & 41.04\\
Ours w/o domain loss & 43.79 &    11.26 & 47.87 & 1.27 & 0.18 & 0.90 & 0.03 & 0.07 & 49.93 & 0.27 & 51.64 & 3.84 & 0.00 & 17.87 & 0.11 & 0.00    & 0.00 & 0.00 & 0.19 & 25.62 & 12.74 & 54.65\\
Ours & \textbf{80.34} &    \textbf{31.82} &    \textbf{58.62} &    2.11 &    \textbf{2.84} &    \textbf{7.66} &    \textbf{1.10} &    \textbf{2.59} &    \textbf{64.08} &    \textbf{10.19} &    \textbf{58.37} &    \textbf{10.70} &    0.00 &    \textbf{51.22} &    0.20 & \textbf{1.55} &    \textbf{0.06} &    \textbf{0.36} &    \textbf{6.63} &    \textbf{59.72} & \textbf{21.71} & \textbf{77.39} \\
\hline
Target (Oracle) & 87.84 & 47.23 & 67.23 & 2.80 &    2.46 &    14.32 &    3.34 &    11.41 &    71.94 &    11.44 &    63.69 &    14.56 &    0.00 & 60.41 & 0.82 & 2.02 & 0.14 & 0.47 &    9.79 & 60.39 & 26.61 & 84.27\\
\hline
\end{tabular}
\end{adjustbox}
\end{table}

\subsubsection{Real-to-real unsupervised image translation}
\flushleft{\textbf{Dataset}}
Following~\cite{zhu_unpaired_2017,alami_mejjati_unsupervised_2018} we show the \textit{real-to-real} image-to-image translation capabilities of our network on images extracted from the COCO~\cite{lin2014microsoft} dataset. Namely, we choose to translate horses to zebras and sheep to giraffes. 

\textbf{Evaluation Protocol} For Quantitative evaluation, we compare the visual similarity between the source and target distributions using the Frechet Inception distance~\cite{heusel2017gans}. Additionally, we show a few qualitative samples, comparing the different baselines.

\begin{figure}[!hbpt]
    \centering
    \def\svgwidth{\columnwidth}
    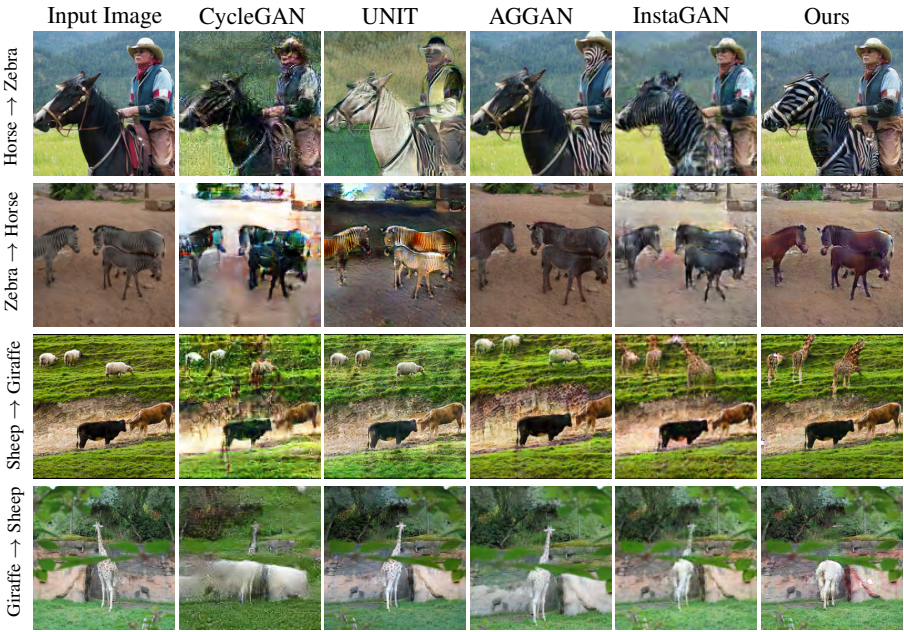
    \caption{\textbf{Visual results for unsupervised image-to-image translation}. We show sample results for the \textit{real-to-real} image-to-image translation task on images extracted from the COCO dataset. Results include Horse $\to$ Zebra (top row), Zebra $\to$ Horse (second row), Sheep $\to$ Giraffe (third row) and Giraffe $\to$ Sheep (bottom row).}
    \label{fig:animals_pre}
\end{figure}

\textbf{Results} We compare our approach with the state-of-the-art methods~\cite{zhu_unpaired_2017,liu2017unsupervised,li_semantic-aware_2018,mo_instance-aware_2019}. Table~\ref{tab:animals} presents the quantitative comparison of visual similarity between the source and the target distribution. The results show that our method performs favorably in both translation directions, from horse to zebra, and zebra to horse, outperforming the other approaches by a large margin. Figure~\ref{fig:animals_pre} shows a few qualitative examples of the proposed translation. The qualitative examples show that our network performs favorably in the texture mapping cases (top two rows) as well as in the case where large geometric changes between foreground objects are necessary.

\begin{table}[htbp!]
\caption{\textbf{Quantitative results of the real-to-real unsupervised image-to-image translation}. We report visual similarity on the distribution level using the Frechet Inception Distance~\cite{heusel2017gans} (lower is better). We mark the best approach in \textbf{bold}.} \label{tab:animals}
\centering
\begin{tabular}{l|c|c}
\hline
\textbf{Method} & \textbf{Horse to Zebra} & \textbf{Zebra to Horse}\\
\hline
CycleGAN~\cite{zhu_unpaired_2017} & 238.41 & 201.65 \\
UNIT~\cite{liu2017unsupervised} & 251.44  & 213.99\\
AG-GAN~\cite{alami_mejjati_unsupervised_2018} & 199.21 & 195.38\\
InstaGAN~\cite{mo_instance-aware_2019} & 220.26 & 209.20\\
Ours & \textbf{104.57} & \textbf{190.01} \\
\hline
\end{tabular}
\end{table}

\subsection{Object Transfiguration}
In object transfiguration, we translate two images which contain foreground objects with different geometry. We present experimental results on a new shape transfiguration dataset that visualizes the difficulties of this task.

\textbf{Dataset} We introduce a new dataset for controlled object transfiguration. We sample random images from the COCO~\cite{lin2014microsoft} dataset and superimpose them with simple geometric shapes, such as circles, triangles, and squares. In total, we create six object transfiguration scenarios: translating circles to triangles, squares to circles, squares to triangles, and their reverse mappings. By abstracting the geometric shapes and randomizing the background, we create a constrained dataset to evaluate the network's capability to translate objects. 

\textbf{Evaluation Protocol} To evaluate the network's ability to transform geometric shapes, we conduct a user ``preference'' study. We gathered data from 10 participants for each of the six transformation scenarios. Participants were shown 50 samples of the baseline results (samples are chosen at random) and had to choose the result that matched the task most closely. We report the average preference score for each of the six transformation scenarios. 
While perceptual studies may capture human visual preference, we also seek an automatic quantitative measure that does not require human trials. We evaluate all methods based on the visual similarity to the target distribution using the Frechet Inception distance~\cite{heusel2017gans}, the pixel difference between images in terms of the $L_1$ norm, and the structural similarity index (SSIM). We compute these metrics on the full images and images masked with the ground truth target objects. The comparison with the masked images captures the network's ability to change the foreground objects, without considering background changes.

\textbf{Results} We compare our approach with CycleGAN~\cite{zhu_unpaired_2017}, UNIT~\cite{liu2017unsupervised}, AG-GAN~\cite{alami_mejjati_unsupervised_2018} and InstaGAN~\cite{mo_instance-aware_2019}. Table~\ref{tab:user_study} shows the user preference study results. In the user study, participants preferred our method in five out of six cases. 

\begin{table}[htbp!]
\caption{\textbf{Experimental results of our user preference study for object transfiguration.} We present results for all six object transfiguration scenarios.} \label{tab:user_study}
\begin{adjustbox}{width=1\textwidth}
\centering
\begin{tabular}{l|c|c|c|c|c|c}
\hline
\multicolumn{7}{c}{\textbf{Shapes Dataset - User Study}}\\
\hline
\textbf{Method} &\textbf{circle to square}&  \textbf{circle to triangle} & \textbf{square to circle} & \textbf{square to triangle} & \textbf{triangle to circle} & \textbf{triangle to square}\\
\hline
CycleGAN~\cite{zhu_unpaired_2017} & 6.86\% & 7.43\% & 3.54\% & 1.27\% & 2.73\% & 4.73\%\\
UNIT~\cite{liu2017unsupervised} & 0.54\% &    0.00\% & 0.36\% & 0.18\% &     0.18\%    & 0.00\%  \\
AG-GAN~\cite{alami_mejjati_unsupervised_2018} & 3.25\% & 3.63\% &    5.19\% & 8.00\% & 5.27\%    & 4.54\%\\
InstaGAN~\cite{mo_instance-aware_2019} & \textbf{57.58\%} & 22.64\% &    38.00\%    & 2.91\%    & 2.55\% & 2.73\%\\
Ours & 31.77\%    & \textbf{66.30\%} & \textbf{    52.91\%} &    \textbf{87.64\%} &    \textbf{89.27\%} &     \textbf{88.00\%}\\
\hline
\end{tabular}
\end{adjustbox}
\end{table}

The top block of Table~\ref{tab:shapes} presents the quantitative results evaluated on the whole image, while the bottom block compares the masked images. Our method and InstaGAN each perform best on three out of the six datasets when we compare FID on the full images. Additionally, AG-GAN and InstaGAN both perform well in terms of $L_1$ and SSIM metrics. Both AG-GAN and InstaGAN mask out the foreground object and apply the transformation solely on the masked content, while our network transforms the whole image. Comparing $L_1$ norm and SSIM on the full image, therefore, gives AG-GAN and InstaGAN an advantage. In contrast, our network makes small changes to the background, which have a strong influence on the $L_1$ and SSIM metrics.
For the masked images, our network outperforms the baselines in four out of six cases on FID and is competitive in terms of $L_1$ and SSIM metrics.

\begin{table}[htbp!]
\caption{\textbf{Quantitative results for the object transfiguration task on the shapes datset}. We report distance between the distributions measured by FID (lower is better), $L_1$ norm between images (lower is better) and the SSIM (higher is better).} \label{tab:shapes}
\begin{adjustbox}{width=1\textwidth}
\centering
\begin{tabular}{ll|ccc|ccc|ccc|ccc|ccc|ccc}
\hline
\multicolumn{20}{c}{\textbf{Shapes Dataset - Quantitative}}\\
\hline
\multicolumn{2}{c|}{\multirow{2}{*}{\textbf{Method}}} & \multicolumn{3}{c|}{\textbf{circle to triangle}} & \multicolumn{3}{c|}{\textbf{square to circle}} & \multicolumn{3}{c|}{\textbf{square to triangle}} & \multicolumn{3}{c|}{\textbf{triangle to circle}} & 
\multicolumn{3}{c|}{\textbf{circle to square}} & \multicolumn{3}{c}{\textbf{triangle to square}}\\
& & FID & L1 & SSIM & FID & L1 & SSIM & FID & L1  & SSIM & FID & L1  & SSIM & FID & L1  & SSIM & FID & L1  & SSIM\\ \hline
\multirow{5}{*}{\rotatebox{90}{Full Image}} & CycleGAN~\cite{zhu_unpaired_2017} & 178.73 & 0.113 & 0.53
& 171.66 & 0.11 & 0.51 
& 143.20 & 0.12 & 0.46 
& 173.81 & 0.12 & 0.47
& 179.84 & 0.10 & 0.60
& 137.97 & 0.11 & 0.55 \\
& UNIT~\cite{liu2017unsupervised} & 219.08 & 0.18 & 0.38 
& 172.00 & 0.37 & 0.09
& 201.69 & 0.18 & 0.34
& 186.65 & 0.17 & 0.43 
& 226.8 & 0.36 & 0.11
& 219.63 & 0.19 & 0.36 \\
& AG-GAN~\cite{alami_mejjati_unsupervised_2018} & 92.72 & 0.05 & 0.8
& 125.26 & 0.04 & 0.85
& 81.34 & 0.05 & 0.81 
& 115.59 & 0.05 & 0.81 
& 85.68 & 0.03 & 0.86 
& 124.25 & 0.05 & 0.82\\
& InstaGAN~\cite{mo_instance-aware_2019} & \textbf{56.03} & 0.079 & 0.73
& \textbf{59.46} & 0.07 & 0.79
& 119.52 & 0.14 & 0.35
& 94.53 & 0.08 & 0.75 
& \textbf{78.17} & 0.07 & 0.76 
& 110.90 & 0.10 & 0.60\\
& Ours & 73.6 & 0.09 & 0.69
& 81.25 & 0.08 & 0.73
& \textbf{69.34} & 0.09 & 0.7
& \textbf{66.53} & 0.07 & 0.77
& 88.81 & 0.08 & 0.77
& \textbf{90.32} & 0.08 & 0.76\\

\hline

\multirow{5}{*}{\rotatebox{90}{Foreground}} & CycleGAN~\cite{zhu_unpaired_2017} & 77.63 & 0.02 & 0.89
& 95.66 & 0.02 & 0.91 
& 71.21 & 0.02 & 0.89
& 134.61 & 0.04 & 0.83
& 252.46 & 0.04 & 0.83
& 126.06 & 0.05 & 0.80 \\
& UNIT~\cite{liu2017unsupervised} & 112.16 & 0.05 & 0.86 
& 101.21 & 0.06 & 0.87
& 71.63 & 0.05 & 0.88
& 216.19 & 0.06 & 0.81 
& 226.84 & 0.08 & 0.79
& 120.91 & 0.09 & 0.79 \\
& AG-GAN~\cite{alami_mejjati_unsupervised_2018} & 68.32 & 0.02 & 0.93
& 104.91 & 0.01 & 0.95
& 44.28 & 0.01 & 0.95
& 168.73 & 0.03 & 0.88
& 158.31 & 0.02 & 0.89 
& 199.44 & 0.07 & 0.75\\
& InstaGAN~\cite{mo_instance-aware_2019} & 57.82 & 0.02 & 0.92
& \textbf{46.07} & 0.01 & 0.95
& 80.97 & 0.02 & 0.85
& 109.93 & 0.03 & 0.87
& \textbf{129.34} & 0.03 & 0.87 
& 124.11 & 0.05 & 0.79\\
& Ours & \textbf{51.04} & 0.02 & 0.92
& 57.36 & 0.02 & 0.95
& \textbf{33.37} & 0.02 & 0.94
& \textbf{59.77} & 0.02 & 0.93
& 133.94 & 0.03 & 0.90
& \textbf{101.22} & 0.03 & 0.91\\
\hline
\end{tabular}
\end{adjustbox}
\end{table}

\begin{figure}[!hbpt]
    \centering
    \def\svgwidth{\columnwidth}
    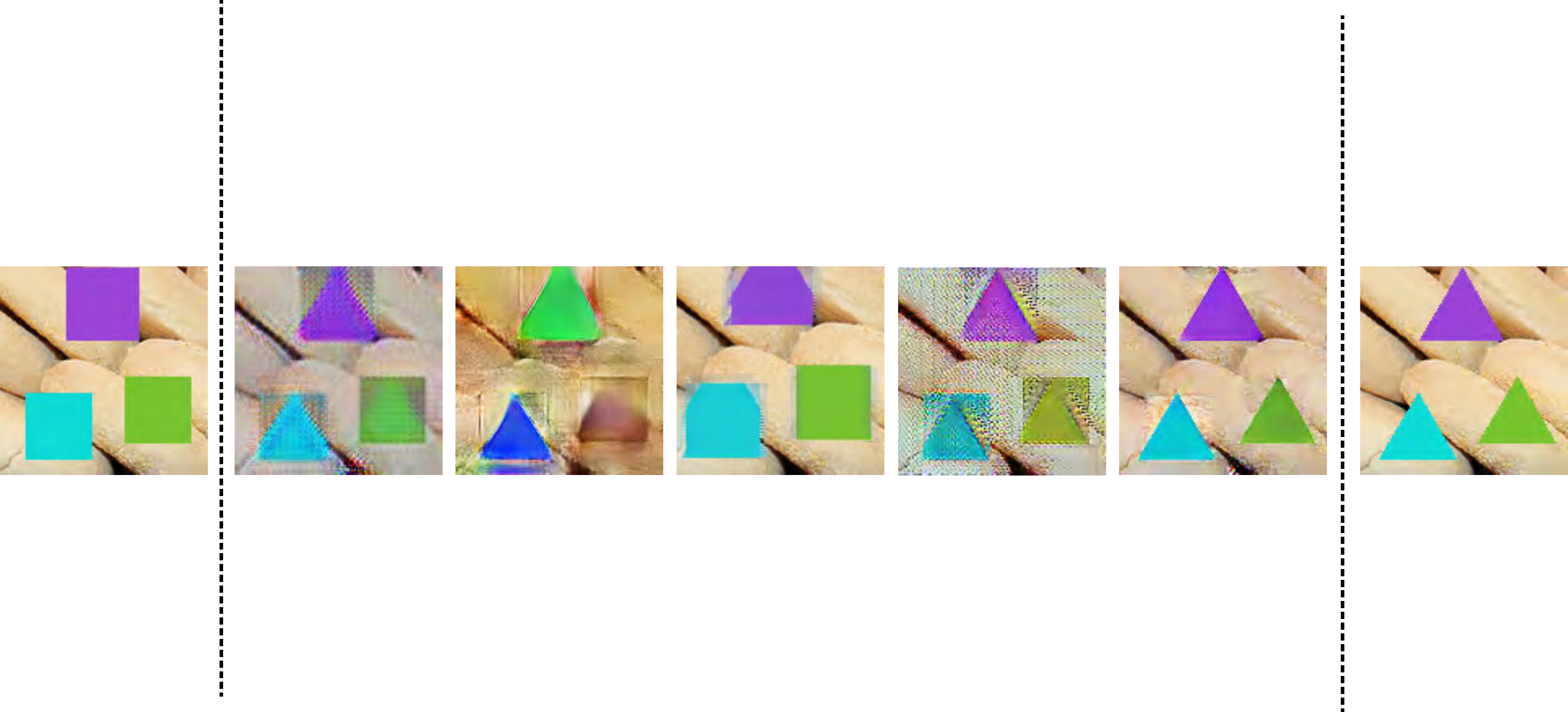
    \caption{\textbf{Visual results for geometric object transfiguration}. We use our shapes dataset to translate from circles to squares (top row), squares to triangles (middle row), and triangles to circles (bottom row). For each sample image, we present our results in comparison with the state-of-the-art baselines.}
    \label{fig:shapes_qual}
\end{figure}

Figure~\ref{fig:shapes_qual} presents examples that demonstrate the effectiveness of our approach. Our network produces sharper boundaries of the objects while preserving the background structure. InstaGAN and UNIT fail to preserve the boundaries and the color structure within the foreground object. CycleGAN and AG-GAN fail to transform the geometry of the foreground object completely and produce outputs close to the source domain.

\begin{figure}[!hbpt]
    \centering
    \includegraphics[width=\textwidth]{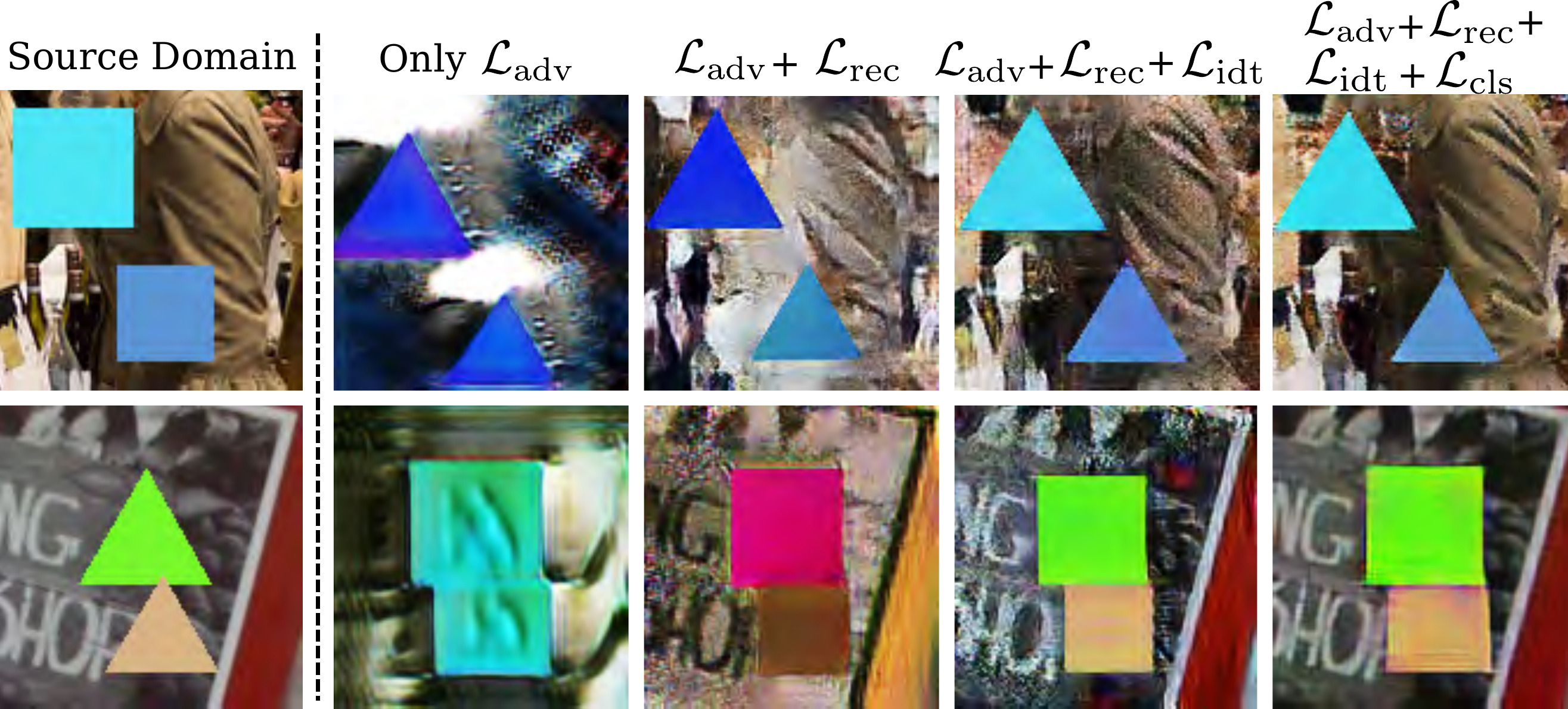}
    \caption{\textbf{Visual results of the objective function ablation}. We present the source image and results for the step-wise ablation of the loss.}
    \label{fig:shapes_geom_ablation}
\end{figure}

In Figure~\ref{fig:shapes_geom_ablation}, we present qualitative comparisons of ablations of our full objective. Removing the class preserving loss decreases the preservation of the background, and the foreground object boundaries become diluted. Removing the identity loss leaves the network to change the color of the objects and background, and removing the cyclic consistency leads to color artifacts in the foreground and background. 

\subsection{Limitations}\label{subsec:limitations}
\begin{figure}[!hbpt]
    \centering
    \def\svgwidth{\columnwidth}
    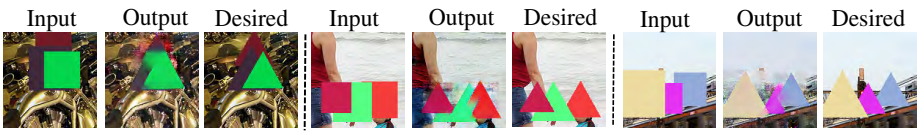
    \caption{Failure Cases. (a) Wrong foreground color (b) Too many triangles (c) Incorrect depth ordering}
    \label{fig:limitations}
\end{figure}

Due to lack of instance level information,  our network may encounter issues in the object transfiguration task due to: - Failure to disentangle the foreground colors of overlapping objects (Figure ~\ref{fig:limitations}(a)). - Replacement of the source objects with more/less target objects (Figure~\ref{fig:limitations}(b)). - Wrong depth ordering of translated objects (Figure~\ref{fig:limitations}(c)).

%% file: gta2city_qualitative.pdf_tex
\begingroup%
  \makeatletter%
  \providecommand\color[2][]{%
    \errmessage{(Inkscape) Color is used for the text in Inkscape, but the package 'color.sty' is not loaded}%
    \renewcommand\color[2][]{}%
  }%
  \providecommand\transparent[1]{%
    \errmessage{(Inkscape) Transparency is used (non-zero) for the text in Inkscape, but the package 'transparent.sty' is not loaded}%
    \renewcommand\transparent[1]{}%
  }%
  \providecommand\rotatebox[2]{#2}%
  \newcommand*\fsize{\dimexpr\f@size pt\relax}%
  \newcommand*\lineheight[1]{\fontsize{\fsize}{#1\fsize}\selectfont}%
  \ifx\svgwidth\undefined%
    \setlength{\unitlength}{1779.06126969bp}%
    \ifx\svgscale\undefined%
      \relax%
    \else%
      \setlength{\unitlength}{\unitlength * \real{\svgscale}}%
    \fi%
  \else%
    \setlength{\unitlength}{\svgwidth}%
  \fi%
  \global\let\svgwidth\undefined%
  \global\let\svgscale\undefined%
  \makeatother%
  \begin{picture}(1,0.37755567)%
    \lineheight{1}%
    \setlength\tabcolsep{0pt}%
    \put(0.72543522,0.37004644){\color[rgb]{0,0,0}\makebox(0,0)[lt]{\lineheight{1.25}\smash{\begin{tabular}[t]{l}Ours\end{tabular}}}}%
    \put(0.84346673,0.36912249){\color[rgb]{0,0,0}\makebox(0,0)[lt]{\lineheight{1.25}\smash{\begin{tabular}[t]{l}Target Domain\end{tabular}}}}%
    \put(0.53260166,0.3690531){\color[rgb]{0,0,0}\makebox(0,0)[lt]{\lineheight{1.25}\smash{\begin{tabular}[t]{l}SG-GAN\end{tabular}}}}%
    \put(-0.00084973,0.36961092){\color[rgb]{0,0,0}\makebox(0,0)[lt]{\lineheight{1.25}\smash{\begin{tabular}[t]{l}Source Domain\end{tabular}}}}%
    \put(0.38372503,0.36929336){\color[rgb]{0,0,0}\makebox(0,0)[lt]{\lineheight{1.25}\smash{\begin{tabular}[t]{l}UNIT\end{tabular}}}}%
    \put(0.20026763,0.36928449){\color[rgb]{0,0,0}\makebox(0,0)[lt]{\lineheight{1.25}\smash{\begin{tabular}[t]{l}CycleGAN\end{tabular}}}}%
    \put(0,0){\includegraphics[width=\unitlength,page=1]{gta2city_qualitative.pdf}}%
    \put(0.76545859,0.07728067){\color[rgb]{0,0,0}\makebox(0,0)[lt]{\lineheight{1.25000012}\smash{\begin{tabular}[t]{l}Ours\end{tabular}}}}%
    \put(0.86207715,0.07728067){\color[rgb]{0,0,0}\makebox(0,0)[lt]{\lineheight{1.25000012}\smash{\begin{tabular}[t]{l}Ground Truth\end{tabular}}}}%
    \put(0.6018726,0.07813222){\color[rgb]{0,0,0}\makebox(0,0)[lt]{\lineheight{1.25000012}\smash{\begin{tabular}[t]{l}SG-GAN\end{tabular}}}}%
    \put(0.05135844,0.07937446){\color[rgb]{0,0,0}\makebox(0,0)[lt]{\lineheight{1.25}\smash{\begin{tabular}[t]{l}Input\end{tabular}}}}%
    \put(0.46730527,0.07813719){\color[rgb]{0,0,0}\makebox(0,0)[lt]{\lineheight{1.25000012}\smash{\begin{tabular}[t]{l}UNIT\end{tabular}}}}%
    \put(0.30299932,0.07939382){\color[rgb]{0,0,0}\makebox(0,0)[lt]{\lineheight{1.25}\smash{\begin{tabular}[t]{l}CycleGAN\end{tabular}}}}%
    \put(0,0){\includegraphics[width=\unitlength,page=2]{gta2city_qualitative.pdf}}%
    \put(0.15603952,0.07939382){\color[rgb]{0,0,0}\makebox(0,0)[lt]{\lineheight{1.25}\smash{\begin{tabular}[t]{l}Source Only\end{tabular}}}}%
    \put(0,0){\includegraphics[width=\unitlength,page=3]{gta2city_qualitative.pdf}}%
  \end{picture}%
\endgroup%

%% file: animals.pdf_tex
\begingroup%
  \makeatletter%
  \providecommand\color[2][]{%
    \errmessage{(Inkscape) Color is used for the text in Inkscape, but the package 'color.sty' is not loaded}%
    \renewcommand\color[2][]{}%
  }%
  \providecommand\transparent[1]{%
    \errmessage{(Inkscape) Transparency is used (non-zero) for the text in Inkscape, but the package 'transparent.sty' is not loaded}%
    \renewcommand\transparent[1]{}%
  }%
  \providecommand\rotatebox[2]{#2}%
  \newcommand*\fsize{\dimexpr\f@size pt\relax}%
  \newcommand*\lineheight[1]{\fontsize{\fsize}{#1\fsize}\selectfont}%
  \ifx\svgwidth\undefined%
    \setlength{\unitlength}{621.05518251bp}%
    \ifx\svgscale\undefined%
      \relax%
    \else%
      \setlength{\unitlength}{\unitlength * \real{\svgscale}}%
    \fi%
  \else%
    \setlength{\unitlength}{\svgwidth}%
  \fi%
  \global\let\svgwidth\undefined%
  \global\let\svgscale\undefined%
  \makeatother%
  \begin{picture}(1,0.67563049)%
    \lineheight{1}%
    \setlength\tabcolsep{0pt}%
    \put(0,0){\includegraphics[width=\unitlength,page=1]{animals.pdf}}%
    \put(0.06314985,0.65268964){\color[rgb]{0,0,0}\makebox(0,0)[lt]{\lineheight{1.25}\smash{\begin{tabular}[t]{l}Input Image\end{tabular}}}}%
    \put(0.22951553,0.65243059){\color[rgb]{0,0,0}\makebox(0,0)[lt]{\lineheight{1.25}\smash{\begin{tabular}[t]{l}CycleGAN\end{tabular}}}}%
    \put(0.40787216,0.65233304){\color[rgb]{0,0,0}\makebox(0,0)[lt]{\lineheight{1.25}\smash{\begin{tabular}[t]{l}UNIT\end{tabular}}}}%
    \put(0.54708589,0.65155099){\color[rgb]{0,0,0}\makebox(0,0)[lt]{\lineheight{1.25}\smash{\begin{tabular}[t]{l}AGGAN\end{tabular}}}}%
    \put(0.70468027,0.65268657){\color[rgb]{0,0,0}\makebox(0,0)[lt]{\lineheight{1.25}\smash{\begin{tabular}[t]{l}InstaGAN\end{tabular}}}}%
    \put(0.89246374,0.65166693){\color[rgb]{0,0,0}\makebox(0,0)[lt]{\lineheight{1.25}\smash{\begin{tabular}[t]{l}Ours\end{tabular}}}}%
    \put(0.0292276,0.49876436){\color[rgb]{0,0,0}\rotatebox{89.831077}{\makebox(0,0)[lt]{\lineheight{1.25}\smash{\begin{tabular}[t]{l}\scriptsize Horse $\to$ Zebra\end{tabular}}}}}%
    \put(0.03002373,0.34308706){\color[rgb]{0,0,0}\rotatebox{89.831077}{\makebox(0,0)[lt]{\lineheight{1.25}\smash{\begin{tabular}[t]{l}\scriptsize Zebra $\to$ Horse\end{tabular}}}}}%
    \put(0.03254698,0.17134848){\color[rgb]{0,0,0}\rotatebox{89.831077}{\makebox(0,0)[lt]{\lineheight{1.25}\smash{\begin{tabular}[t]{l}\scriptsize Sheep  $\to$ Giraffe\end{tabular}}}}}%
    \put(0.03306855,0.01376391){\color[rgb]{0,0,0}\rotatebox{89.831077}{\makebox(0,0)[lt]{\lineheight{1.25}\smash{\begin{tabular}[t]{l}\scriptsize Giraffe  $\to$ Sheep\end{tabular}}}}}%
  \end{picture}%
\endgroup%

%% file: shapes_qual.pdf_tex
\begingroup%
  \makeatletter%
  \providecommand\color[2][]{%
    \errmessage{(Inkscape) Color is used for the text in Inkscape, but the package 'color.sty' is not loaded}%
    \renewcommand\color[2][]{}%
  }%
  \providecommand\transparent[1]{%
    \errmessage{(Inkscape) Transparency is used (non-zero) for the text in Inkscape, but the package 'transparent.sty' is not loaded}%
    \renewcommand\transparent[1]{}%
  }%
  \providecommand\rotatebox[2]{#2}%
  \newcommand*\fsize{\dimexpr\f@size pt\relax}%
  \newcommand*\lineheight[1]{\fontsize{\fsize}{#1\fsize}\selectfont}%
  \ifx\svgwidth\undefined%
    \setlength{\unitlength}{878.38128662bp}%
    \ifx\svgscale\undefined%
      \relax%
    \else%
      \setlength{\unitlength}{\unitlength * \real{\svgscale}}%
    \fi%
  \else%
    \setlength{\unitlength}{\svgwidth}%
  \fi%
  \global\let\svgwidth\undefined%
  \global\let\svgscale\undefined%
  \makeatother%
  \begin{picture}(1,0.45409779)%
    \lineheight{1}%
    \setlength\tabcolsep{0pt}%
    \put(0,0){\includegraphics[width=\unitlength,page=1]{shapes_qual.pdf}}%
    \put(0.04303804,0.43879066){\color[rgb]{0,0,0}\makebox(0,0)[lt]{\lineheight{1.25}\smash{\begin{tabular}[t]{l}Input\end{tabular}}}}%
    \put(0.15545925,0.4387909){\color[rgb]{0,0,0}\makebox(0,0)[lt]{\lineheight{1.25}\smash{\begin{tabular}[t]{l}CycleGAN\end{tabular}}}}%
    \put(0.32710193,0.43624938){\color[rgb]{0,0,0}\makebox(0,0)[lt]{\lineheight{1.25}\smash{\begin{tabular}[t]{l}UNIT\end{tabular}}}}%
    \put(0.45286239,0.43623604){\color[rgb]{0,0,0}\makebox(0,0)[lt]{\lineheight{1.25}\smash{\begin{tabular}[t]{l}AGGAN\end{tabular}}}}%
    \put(0.58825808,0.43624938){\color[rgb]{0,0,0}\makebox(0,0)[lt]{\lineheight{1.25}\smash{\begin{tabular}[t]{l}InstaGAN\end{tabular}}}}%
    \put(0.76047893,0.43624938){\color[rgb]{0,0,0}\makebox(0,0)[lt]{\lineheight{1.25}\smash{\begin{tabular}[t]{l}Ours\end{tabular}}}}%
    \put(0.86684294,0.43624938){\color[rgb]{0,0,0}\makebox(0,0)[lt]{\lineheight{1.25}\smash{\begin{tabular}[t]{l}Ground Truth\end{tabular}}}}%
    \put(0,0){\includegraphics[width=\unitlength,page=2]{shapes_qual.pdf}}%
  \end{picture}%
\endgroup%

%% file: failure_cases.pdf_tex
\begingroup%
  \makeatletter%
  \providecommand\color[2][]{%
    \errmessage{(Inkscape) Color is used for the text in Inkscape, but the package 'color.sty' is not loaded}%
    \renewcommand\color[2][]{}%
  }%
  \providecommand\transparent[1]{%
    \errmessage{(Inkscape) Transparency is used (non-zero) for the text in Inkscape, but the package 'transparent.sty' is not loaded}%
    \renewcommand\transparent[1]{}%
  }%
  \providecommand\rotatebox[2]{#2}%
  \newcommand*\fsize{\dimexpr\f@size pt\relax}%
  \newcommand*\lineheight[1]{\fontsize{\fsize}{#1\fsize}\selectfont}%
  \ifx\svgwidth\undefined%
    \setlength{\unitlength}{942.88993295bp}%
    \ifx\svgscale\undefined%
      \relax%
    \else%
      \setlength{\unitlength}{\unitlength * \real{\svgscale}}%
    \fi%
  \else%
    \setlength{\unitlength}{\svgwidth}%
  \fi%
  \global\let\svgwidth\undefined%
  \global\let\svgscale\undefined%
  \makeatother%
  \begin{picture}(1,0.1270666)%
    \lineheight{1}%
    \setlength\tabcolsep{0pt}%
    \put(0.02523248,0.11216746){\color[rgb]{0,0,0}\makebox(0,0)[lt]{\lineheight{1.25}\smash{\begin{tabular}[t]{l}Input\end{tabular}}}}%
    \put(0,0){\includegraphics[width=\unitlength,page=1]{failure_cases.pdf}}%
    \put(0.13033343,0.1137832){\color[rgb]{0,0,0}\makebox(0,0)[lt]{\lineheight{1.25}\smash{\begin{tabular}[t]{l}Output\end{tabular}}}}%
    \put(0.23123337,0.11207368){\color[rgb]{0,0,0}\makebox(0,0)[lt]{\lineheight{1.25}\smash{\begin{tabular}[t]{l}Desired\end{tabular}}}}%
    \put(0.35670763,0.11216786){\color[rgb]{0,0,0}\makebox(0,0)[lt]{\lineheight{1.25}\smash{\begin{tabular}[t]{l}Input\end{tabular}}}}%
    \put(0.4658484,0.1137836){\color[rgb]{0,0,0}\makebox(0,0)[lt]{\lineheight{1.25}\smash{\begin{tabular}[t]{l}Output\end{tabular}}}}%
    \put(0.56993094,0.11297593){\color[rgb]{0,0,0}\makebox(0,0)[lt]{\lineheight{1.25}\smash{\begin{tabular}[t]{l}Desired\end{tabular}}}}%
    \put(0.69731215,0.1113602){\color[rgb]{0,0,0}\makebox(0,0)[lt]{\lineheight{1.25}\smash{\begin{tabular}[t]{l}Input\end{tabular}}}}%
    \put(0.80324593,0.11216826){\color[rgb]{0,0,0}\makebox(0,0)[lt]{\lineheight{1.25}\smash{\begin{tabular}[t]{l}Output\end{tabular}}}}%
    \put(0.90732783,0.11216826){\color[rgb]{0,0,0}\makebox(0,0)[lt]{\lineheight{1.25}\smash{\begin{tabular}[t]{l}Desired\end{tabular}}}}%
  \end{picture}%
\endgroup%

%% file: conc.tex
In this work, we presented a generative adversarial network for semantic preserving domain adaptation and object transfiguration. By incorporating the proposed object transfiguration, cross-domain semantic consistency, and joint image/class-map adversarial losses, we improved performance on both tasks over state-of-the-art methods. In object transfiguration, our network translated the geometry consistently and preserved the colors and textures of the objects better than state-of-the-art techniques. As a result, our average human-preference score was higher for five out of the six object transfiguration scenarios. For domain adaptation, our semantic label preserving mechanism led to superior classification performance when using the adapted data to train a semantic segmentation network. Introducing our cross-domain semantic consistency loss improved semantic segmentation performance by 5.4\% mean IoU and 13.5\% pixel accuracy on the GTA5 to Cityscapes domain adaptation task. In the future, we would like to extend our approach to include instance-level information to resolve the current limitations of the object transfiguration tasks (Section~\ref{subsec:limitations}). Another avenue for research is regarding domain sampling. For domain transfer tasks, we sampled images from the target distribution uniformly. In the future, we would like to explore active learning strategies to choose the samples in a more informative way.

%% file: eccv_supplementary.tex
\appendix
\section{Additional Results}
In this section, we present additional qualitative results and comparisons on previously introduced datasets together. 
\subsection{Synthetic-to-Real Domain Adaptation}

\begin{figure*}[h]
    \centering
    \def\svgwidth{\textwidth}
    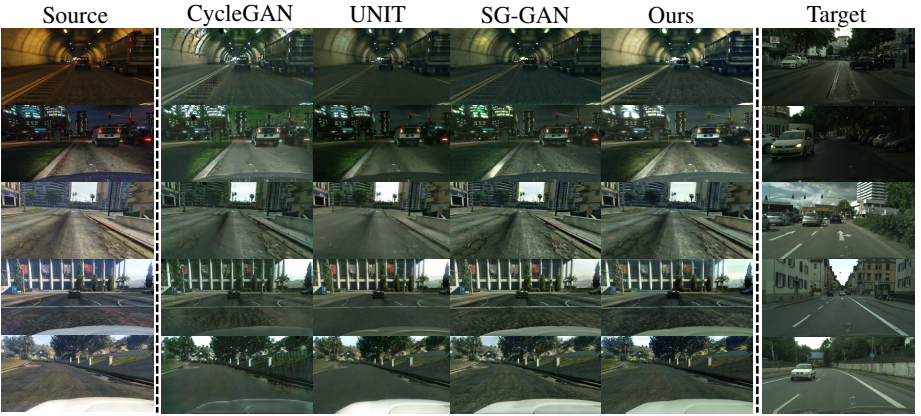
    \caption{\textbf{Additional visual results for synthetic-to-real domain adaptation.} We show additional results of translating the synthetic GTA5 to the real Cityscapes dataset.}
    \label{fig:sup_domain}
\end{figure*}

\begin{figure*}[h]
    \centering
    \def\svgwidth{\textwidth}
    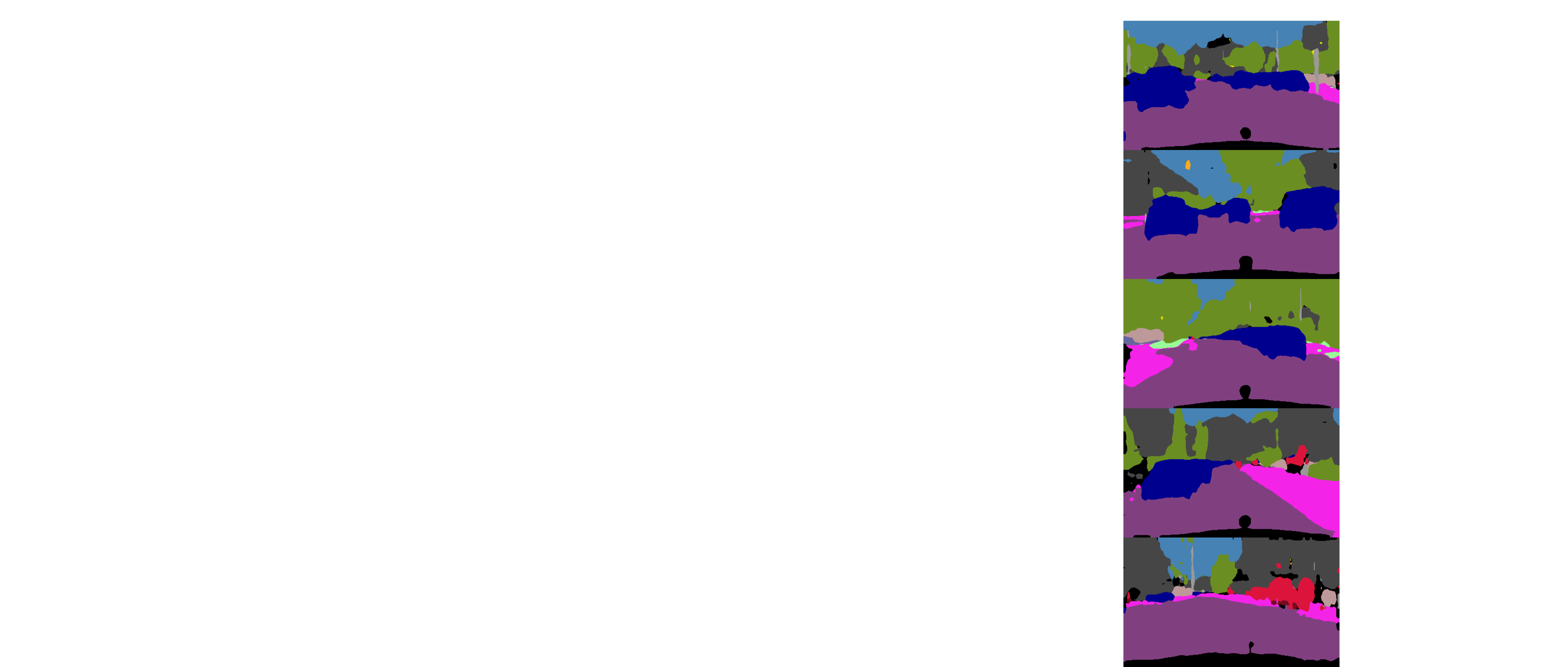
    \caption{\textbf{Additional visual results for synthetic-to-real segmentation performance.} We show additional results of using translated images to train a semantic segmentation network.}
    \label{fig:sup_segm}
\end{figure*}

\subsection{Real-to-real unsupervised image translation}
\begin{figure*}[h]
    \centering
    \def\svgwidth{\textwidth}
    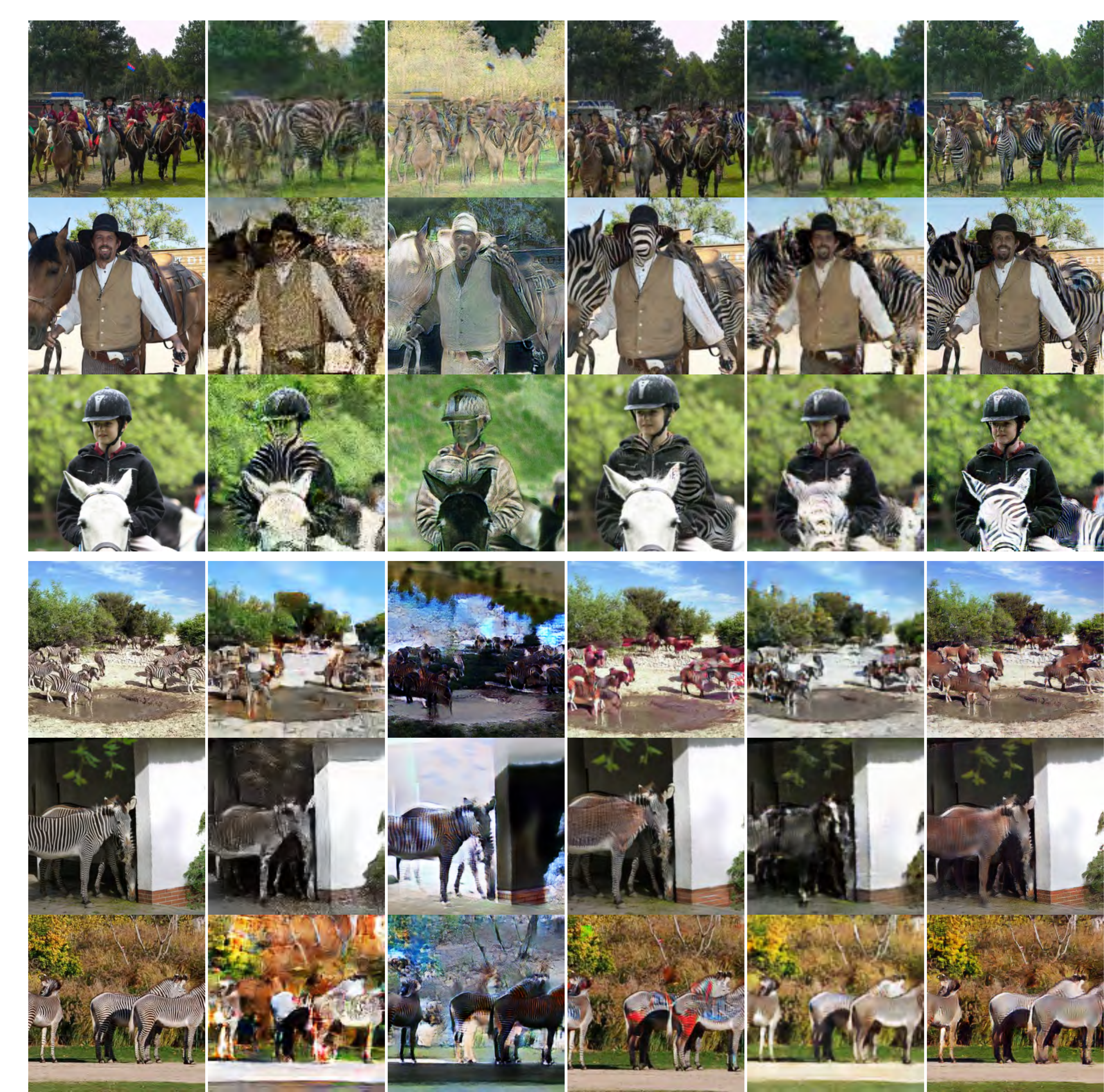
    \caption{\textbf{Additional visual results for unsupervised image-to-image translation}. We show sample results for the \textit{real-to-real} image-to-image translation task on images extracted from the COCO dataset. Results include Horse $\to$ Zebra (top three rows) and Zebra $\to$ Horse (bottom three rows).}
    \label{fig:sup_horse2zebra}
\end{figure*}

\begin{figure*}[h]
    \centering
    \def\svgwidth{\textwidth}
    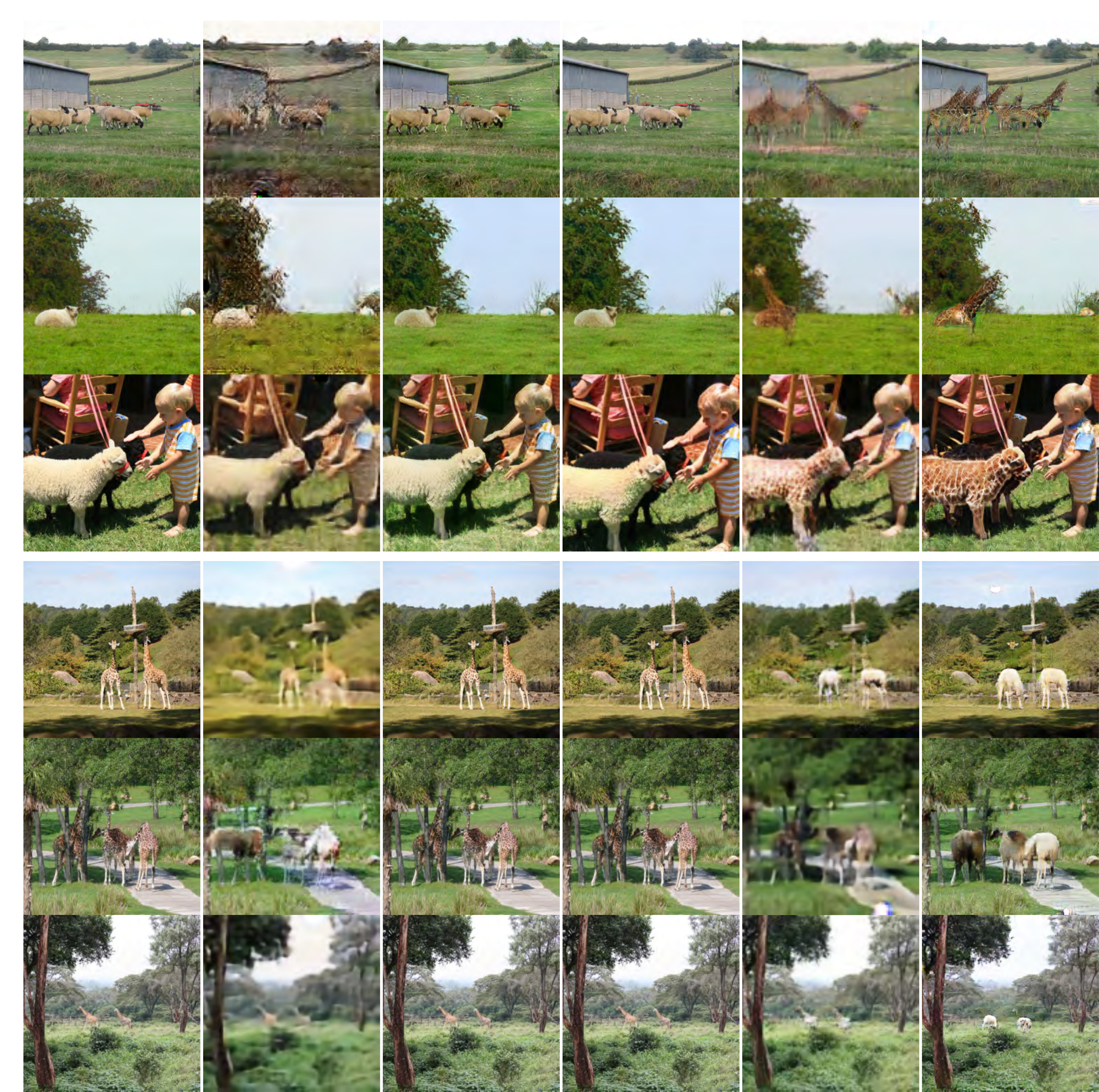
    \caption{\textbf{Additional visual results for unsupervised image-to-image translation}. We show sample results for the \textit{real-to-real} image-to-image translation task on images extracted from the COCO dataset. Results include Sheep $\to$ Giraffe (top three rows) and Giraffe $\to$ Sheep (bottom three rows).}
    \label{fig:sup_sheep2giraffe}
\end{figure*}

Figure~\ref{fig:sup_horse2zebra} and ~\ref{fig:sup_sheep2giraffe}, show a few qualitative examples of our proposed translation. The qualitative examples show that our network performs favorably in the texture mapping cases (Figure~\ref{fig:sup_horse2zebra}) as well as in the case where large geometric changes between foreground objects are necessary. (Figure~\ref{fig:sup_sheep2giraffe}).

\subsection{Object Transfiguration}
\begin{figure*}[h]
    \centering
    \def\svgwidth{\textwidth}
    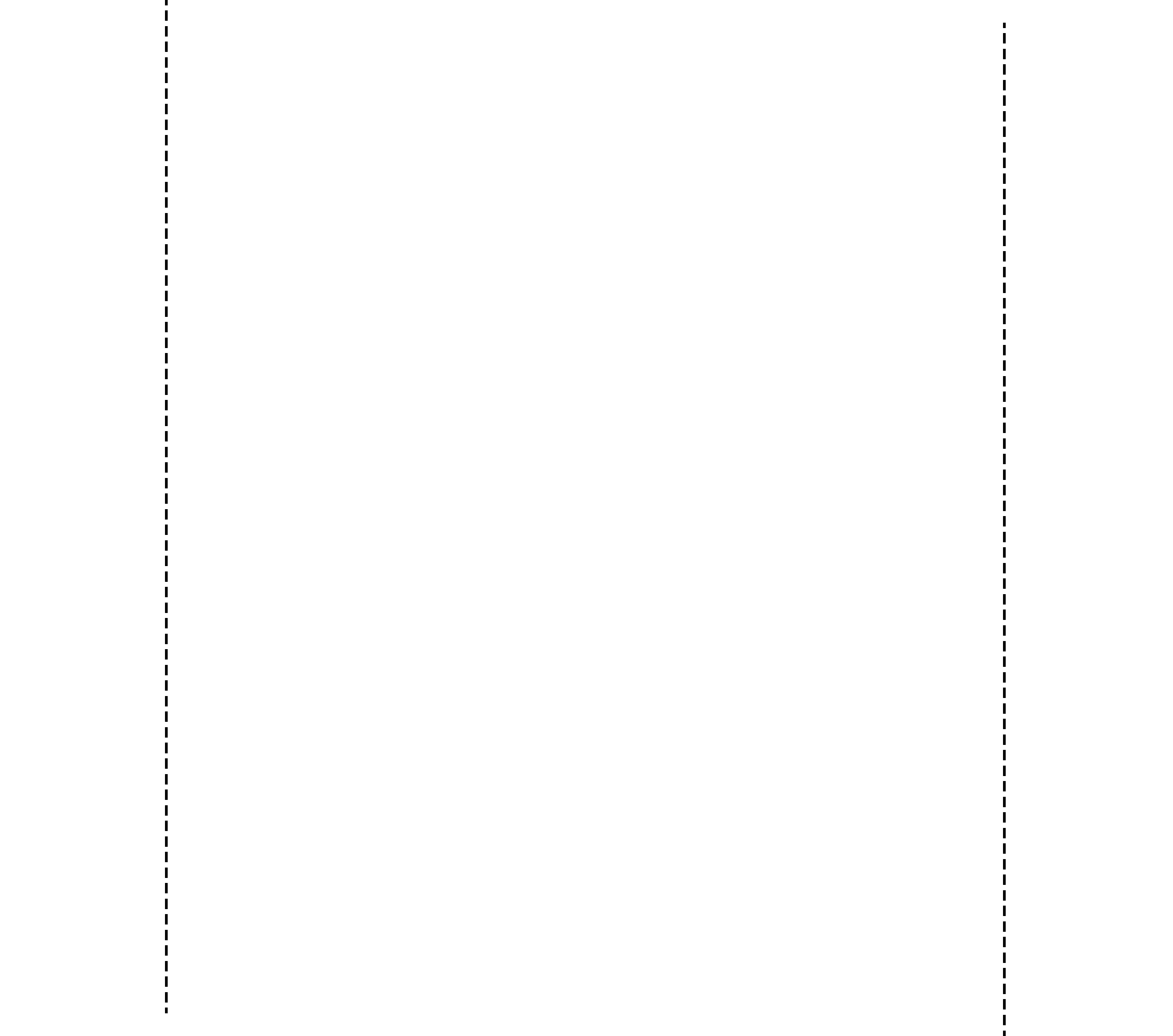
    \caption{\textbf{Additional visual results for geometric object transfiguration}. We use our shapes dataset to translate from triangles to circles (top two rows), circles to squares (middle two rows), and squares to triangles (bottom two rows). For each sample image, we present our results in comparison with the state-of-the-art baselines.}
    \label{fig:sup_shap_qual}
\end{figure*}

\begin{figure*}[h]
    \centering
    \def\svgwidth{\textwidth}
    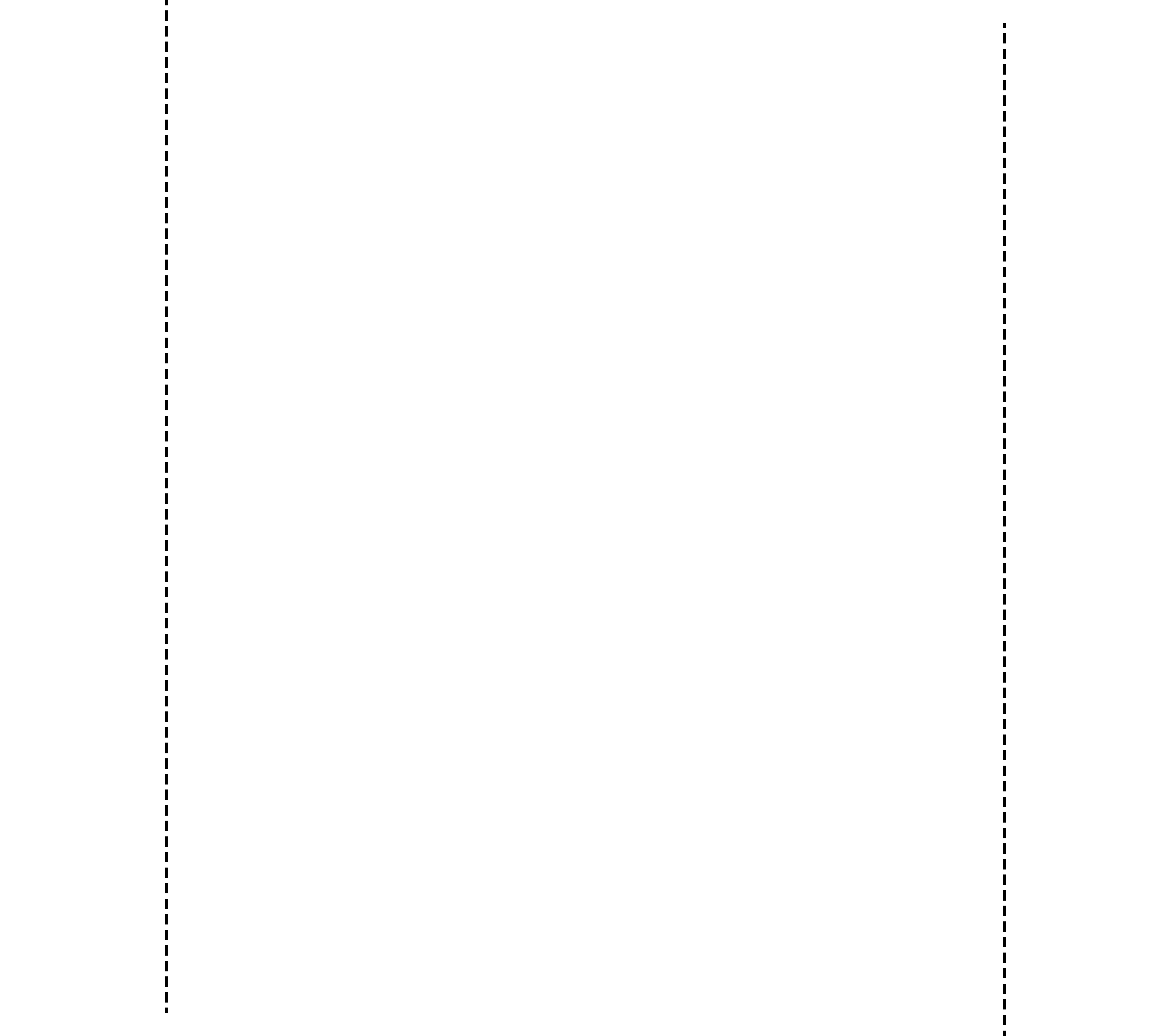
    \caption{\textbf{Additional visual results for geometric object transfiguration}. We use our shapes dataset to translate from circles to triangles (top two rows), triangles to squares (middle two rows), and squares to circles (bottom two rows). For each sample image, we present our results in comparison with the state-of-the-art baselines.}
    \label{fig:sup_shap_qual_reverse}
\end{figure*}

Figure~\ref{fig:sup_shap_qual} and ~\ref{fig:sup_shap_qual_reverse}, present examples that demonstrate the effectiveness of our approach. Our network produces sharper boundaries of the objects while preserving the background structure. InstaGAN and UNIT fail to preserve the boundaries and the color structure within the foreground object. CycleGAN and AG-GAN fail to transform the geometry of the foreground object completely and produce outputs close to the source domain.

\clearpage
\section{Image Resolution}
Due to resource constraints, we relied on down-sampled versions of the original datasets for training. Specifically, we used image size $256 \times 512$ pixels for the GTA5 to Cityscapes experiments, image size $128 \times 128$ pixels for the shape transformation experiments and image size $256 \times 256$ pixels for the anecdotal animal transformation experiments.

\section{Implementation Details}
The discriminators receive concatenate images and semantic maps as input, both normalized similarly to the generator inputs. The generator networks $G_{S\to T}$ and $G_{T\to S}$ and the discriminators $D_{S}$ and $D_{T}$ are all initialized randomly. We use a learning rate of $2E^{-4}$ with momentum $0.9$ and optimize using the Adam optimizer. We trained our model on two NVIDIA Tesla K20X GPUs, each with $12$ GB of memory.  For a detailed description of the network and used parameters for each experiment, please consult the supplementary document.

All networks used in this work were trained on a single machine containing two NVIDIA Tesla K20X GPUs, each with $12$ GB of memory. We used Adam optimizer with the same initial learning rate of $0.0002$. The discriminator was trained with a history of the last $50$ images. We applied instance normalization to both the generator and discriminator. For all the experiments, the networks were trained up to $200$ epochs. For object transfiguration tasks, $\lambda_{dom}$ was set to zero to remove cross-domain consistency. Similarly, for domain translation tasks, $\lambda_{cls}$ was set to zero to turn off class preserving loss. We used $\lambda_{rec} =10, \lambda_{idt}=10$ for all the experiments.

%% file: supplemental_gta_quality.pdf_tex
\begingroup%
  \makeatletter%
  \providecommand\color[2][]{%
    \errmessage{(Inkscape) Color is used for the text in Inkscape, but the package 'color.sty' is not loaded}%
    \renewcommand\color[2][]{}%
  }%
  \providecommand\transparent[1]{%
    \errmessage{(Inkscape) Transparency is used (non-zero) for the text in Inkscape, but the package 'transparent.sty' is not loaded}%
    \renewcommand\transparent[1]{}%
  }%
  \providecommand\rotatebox[2]{#2}%
  \newcommand*\fsize{\dimexpr\f@size pt\relax}%
  \newcommand*\lineheight[1]{\fontsize{\fsize}{#1\fsize}\selectfont}%
  \ifx\svgwidth\undefined%
    \setlength{\unitlength}{1087.50314307bp}%
    \ifx\svgscale\undefined%
      \relax%
    \else%
      \setlength{\unitlength}{\unitlength * \real{\svgscale}}%
    \fi%
  \else%
    \setlength{\unitlength}{\svgwidth}%
  \fi%
  \global\let\svgwidth\undefined%
  \global\let\svgscale\undefined%
  \makeatother%
  \begin{picture}(1,0.43481027)%
    \lineheight{1}%
    \setlength\tabcolsep{0pt}%
    \put(0,0){\includegraphics[width=\unitlength,page=1]{supplemental_gta_quality.pdf}}%
    \put(0.70854124,0.42056003){\color[rgb]{0,0,0}\makebox(0,0)[lt]{\lineheight{1.25}\smash{\begin{tabular}[t]{l}Ours\end{tabular}}}}%
    \put(0,0){\includegraphics[width=\unitlength,page=2]{supplemental_gta_quality.pdf}}%
    \put(0.5248838,0.42063437){\color[rgb]{0,0,0}\makebox(0,0)[lt]{\lineheight{1.25}\smash{\begin{tabular}[t]{l}SG-GAN\end{tabular}}}}%
    \put(0,0){\includegraphics[width=\unitlength,page=3]{supplemental_gta_quality.pdf}}%
    \put(0.38099878,0.42127052){\color[rgb]{0,0,0}\makebox(0,0)[lt]{\lineheight{1.25}\smash{\begin{tabular}[t]{l}UNIT\end{tabular}}}}%
    \put(0,0){\includegraphics[width=\unitlength,page=4]{supplemental_gta_quality.pdf}}%
    \put(0.20324743,0.42223487){\color[rgb]{0,0,0}\makebox(0,0)[lt]{\lineheight{1.25}\smash{\begin{tabular}[t]{l}CycleGAN\end{tabular}}}}%
    \put(0,0){\includegraphics[width=\unitlength,page=5]{supplemental_gta_quality.pdf}}%
    \put(0.04466941,0.42062423){\color[rgb]{0,0,0}\makebox(0,0)[lt]{\lineheight{1.25}\smash{\begin{tabular}[t]{l}Source\end{tabular}}}}%
    \put(0,0){\includegraphics[width=\unitlength,page=6]{supplemental_gta_quality.pdf}}%
    \put(0.88242851,0.42170052){\color[rgb]{0,0,0}\makebox(0,0)[lt]{\lineheight{1.25}\smash{\begin{tabular}[t]{l}Target\end{tabular}}}}%
    \put(0,0){\includegraphics[width=\unitlength,page=7]{supplemental_gta_quality.pdf}}%
  \end{picture}%
\endgroup%

%% file: segmentation_qualitative.pdf_tex
\begingroup%
  \makeatletter%
  \providecommand\color[2][]{%
    \errmessage{(Inkscape) Color is used for the text in Inkscape, but the package 'color.sty' is not loaded}%
    \renewcommand\color[2][]{}%
  }%
  \providecommand\transparent[1]{%
    \errmessage{(Inkscape) Transparency is used (non-zero) for the text in Inkscape, but the package 'transparent.sty' is not loaded}%
    \renewcommand\transparent[1]{}%
  }%
  \providecommand\rotatebox[2]{#2}%
  \newcommand*\fsize{\dimexpr\f@size pt\relax}%
  \newcommand*\lineheight[1]{\fontsize{\fsize}{#1\fsize}\selectfont}%
  \ifx\svgwidth\undefined%
    \setlength{\unitlength}{1740.77590582bp}%
    \ifx\svgscale\undefined%
      \relax%
    \else%
      \setlength{\unitlength}{\unitlength * \real{\svgscale}}%
    \fi%
  \else%
    \setlength{\unitlength}{\svgwidth}%
  \fi%
  \global\let\svgwidth\undefined%
  \global\let\svgscale\undefined%
  \makeatother%
  \begin{picture}(1,0.42515154)%
    \lineheight{1}%
    \setlength\tabcolsep{0pt}%
    \put(0,0){\includegraphics[width=\unitlength,page=1]{segmentation_qualitative.pdf}}%
    \put(0.75891607,0.41699271){\color[rgb]{0,0,0}\makebox(0,0)[lt]{\lineheight{1.25}\smash{\begin{tabular}[t]{l}Ours\end{tabular}}}}%
    \put(0,0){\includegraphics[width=\unitlength,page=2]{segmentation_qualitative.pdf}}%
    \put(0.60336375,0.41614706){\color[rgb]{0,0,0}\makebox(0,0)[lt]{\lineheight{1.25}\smash{\begin{tabular}[t]{l}SG-GAN\\\end{tabular}}}}%
    \put(0,0){\includegraphics[width=\unitlength,page=3]{segmentation_qualitative.pdf}}%
    \put(0.48275807,0.41641851){\color[rgb]{0,0,0}\makebox(0,0)[lt]{\lineheight{1.25}\smash{\begin{tabular}[t]{l}UNIT\end{tabular}}}}%
    \put(0,0){\includegraphics[width=\unitlength,page=4]{segmentation_qualitative.pdf}}%
    \put(0.31391729,0.41614706){\color[rgb]{0,0,0}\makebox(0,0)[lt]{\lineheight{1.25}\smash{\begin{tabular}[t]{l}CycleGAN\\\end{tabular}}}}%
    \put(0.16619312,0.41572424){\color[rgb]{0,0,0}\makebox(0,0)[lt]{\lineheight{1.25}\smash{\begin{tabular}[t]{l}Source Only\end{tabular}}}}%
    \put(0.05191865,0.41675243){\color[rgb]{0,0,0}\makebox(0,0)[lt]{\lineheight{1.25}\smash{\begin{tabular}[t]{l}Input\\\end{tabular}}}}%
    \put(0.85554382,0.41729539){\color[rgb]{0,0,0}\makebox(0,0)[lt]{\lineheight{1.25}\smash{\begin{tabular}[t]{l}Ground Truth\end{tabular}}}}%
    \put(0,0){\includegraphics[width=\unitlength,page=5]{segmentation_qualitative.pdf}}%
  \end{picture}%
\endgroup%

%% file: horse2zebra.pdf_tex
\begingroup%
  \makeatletter%
  \providecommand\color[2][]{%
    \errmessage{(Inkscape) Color is used for the text in Inkscape, but the package 'color.sty' is not loaded}%
    \renewcommand\color[2][]{}%
  }%
  \providecommand\transparent[1]{%
    \errmessage{(Inkscape) Transparency is used (non-zero) for the text in Inkscape, but the package 'transparent.sty' is not loaded}%
    \renewcommand\transparent[1]{}%
  }%
  \providecommand\rotatebox[2]{#2}%
  \newcommand*\fsize{\dimexpr\f@size pt\relax}%
  \newcommand*\lineheight[1]{\fontsize{\fsize}{#1\fsize}\selectfont}%
  \ifx\svgwidth\undefined%
    \setlength{\unitlength}{1197.63295091bp}%
    \ifx\svgscale\undefined%
      \relax%
    \else%
      \setlength{\unitlength}{\unitlength * \real{\svgscale}}%
    \fi%
  \else%
    \setlength{\unitlength}{\svgwidth}%
  \fi%
  \global\let\svgwidth\undefined%
  \global\let\svgscale\undefined%
  \makeatother%
  \begin{picture}(1,0.98935492)%
    \lineheight{1}%
    \setlength\tabcolsep{0pt}%
    \put(0.04887358,0.97904012){\color[rgb]{0,0,0}\makebox(0,0)[lt]{\lineheight{1.25}\smash{\begin{tabular}[t]{l}Input Image\end{tabular}}}}%
    \put(0.21080006,0.97753437){\color[rgb]{0,0,0}\makebox(0,0)[lt]{\lineheight{1.25}\smash{\begin{tabular}[t]{l}CycleGAN\end{tabular}}}}%
    \put(0.39855048,0.97770301){\color[rgb]{0,0,0}\makebox(0,0)[lt]{\lineheight{1.25}\smash{\begin{tabular}[t]{l}UNIT\end{tabular}}}}%
    \put(0.54599569,0.97801117){\color[rgb]{0,0,0}\makebox(0,0)[lt]{\lineheight{1.25}\smash{\begin{tabular}[t]{l}AGGAN\end{tabular}}}}%
    \put(0.708176,0.97917615){\color[rgb]{0,0,0}\makebox(0,0)[lt]{\lineheight{1.25}\smash{\begin{tabular}[t]{l}InstaGAN\end{tabular}}}}%
    \put(0.88029999,0.97668121){\color[rgb]{0,0,0}\makebox(0,0)[lt]{\lineheight{1.25}\smash{\begin{tabular}[t]{l}Ours\end{tabular}}}}%
    \put(0.01674188,0.68241328){\color[rgb]{0,0,0}\rotatebox{89.831077}{\makebox(0,0)[lt]{\lineheight{1.25}\smash{\begin{tabular}[t]{l}Horse $\to$ Zebra\end{tabular}}}}}%
    \put(0.01500679,0.18189191){\color[rgb]{0,0,0}\rotatebox{89.831077}{\makebox(0,0)[lt]{\lineheight{1.25}\smash{\begin{tabular}[t]{l}Zebra $\to$ Horse\end{tabular}}}}}%
    \put(0,0){\includegraphics[width=\unitlength,page=1]{horse2zebra.pdf}}%
  \end{picture}%
\endgroup%

%% file: sheep2giraffe.pdf_tex
\begingroup%
  \makeatletter%
  \providecommand\color[2][]{%
    \errmessage{(Inkscape) Color is used for the text in Inkscape, but the package 'color.sty' is not loaded}%
    \renewcommand\color[2][]{}%
  }%
  \providecommand\transparent[1]{%
    \errmessage{(Inkscape) Transparency is used (non-zero) for the text in Inkscape, but the package 'transparent.sty' is not loaded}%
    \renewcommand\transparent[1]{}%
  }%
  \providecommand\rotatebox[2]{#2}%
  \newcommand*\fsize{\dimexpr\f@size pt\relax}%
  \newcommand*\lineheight[1]{\fontsize{\fsize}{#1\fsize}\selectfont}%
  \ifx\svgwidth\undefined%
    \setlength{\unitlength}{1987.35288124bp}%
    \ifx\svgscale\undefined%
      \relax%
    \else%
      \setlength{\unitlength}{\unitlength * \real{\svgscale}}%
    \fi%
  \else%
    \setlength{\unitlength}{\svgwidth}%
  \fi%
  \global\let\svgwidth\undefined%
  \global\let\svgscale\undefined%
  \makeatother%
  \begin{picture}(1,0.99368706)%
    \lineheight{1}%
    \setlength\tabcolsep{0pt}%
    \put(0.0447089,0.98332709){\color[rgb]{0,0,0}\makebox(0,0)[lt]{\lineheight{1.25}\smash{\begin{tabular}[t]{l}Input Image\end{tabular}}}}%
    \put(0.20734441,0.98181475){\color[rgb]{0,0,0}\makebox(0,0)[lt]{\lineheight{1.25}\smash{\begin{tabular}[t]{l}CycleGAN\end{tabular}}}}%
    \put(0.39591693,0.98198413){\color[rgb]{0,0,0}\makebox(0,0)[lt]{\lineheight{1.25}\smash{\begin{tabular}[t]{l}UNIT\end{tabular}}}}%
    \put(0.54400776,0.98229364){\color[rgb]{0,0,0}\makebox(0,0)[lt]{\lineheight{1.25}\smash{\begin{tabular}[t]{l}AGGAN\end{tabular}}}}%
    \put(0.70689821,0.98346372){\color[rgb]{0,0,0}\makebox(0,0)[lt]{\lineheight{1.25}\smash{\begin{tabular}[t]{l}InstaGAN\end{tabular}}}}%
    \put(0.87977588,0.98095786){\color[rgb]{0,0,0}\makebox(0,0)[lt]{\lineheight{1.25}\smash{\begin{tabular}[t]{l}Ours\end{tabular}}}}%
    \put(0.01243651,0.68540142){\color[rgb]{0,0,0}\rotatebox{89.831077}{\makebox(0,0)[lt]{\lineheight{1.25}\smash{\begin{tabular}[t]{l}Sheep $\to$ Giraffe\end{tabular}}}}}%
    \put(0.01069382,0.18268841){\color[rgb]{0,0,0}\rotatebox{89.831077}{\makebox(0,0)[lt]{\lineheight{1.25}\smash{\begin{tabular}[t]{l}Giraffe $\to$ Sheep\end{tabular}}}}}%
    \put(0,0){\includegraphics[width=\unitlength,page=1]{sheep2giraffe.pdf}}%
  \end{picture}%
\endgroup%

%% file: shapes_supp_qual.pdf_tex
\begingroup%
  \makeatletter%
  \providecommand\color[2][]{%
    \errmessage{(Inkscape) Color is used for the text in Inkscape, but the package 'color.sty' is not loaded}%
    \renewcommand\color[2][]{}%
  }%
  \providecommand\transparent[1]{%
    \errmessage{(Inkscape) Transparency is used (non-zero) for the text in Inkscape, but the package 'transparent.sty' is not loaded}%
    \renewcommand\transparent[1]{}%
  }%
  \providecommand\rotatebox[2]{#2}%
  \newcommand*\fsize{\dimexpr\f@size pt\relax}%
  \newcommand*\lineheight[1]{\fontsize{\fsize}{#1\fsize}\selectfont}%
  \ifx\svgwidth\undefined%
    \setlength{\unitlength}{878.38128662bp}%
    \ifx\svgscale\undefined%
      \relax%
    \else%
      \setlength{\unitlength}{\unitlength * \real{\svgscale}}%
    \fi%
  \else%
    \setlength{\unitlength}{\svgwidth}%
  \fi%
  \global\let\svgwidth\undefined%
  \global\let\svgscale\undefined%
  \makeatother%
  \begin{picture}(1,0.8832117)%
    \lineheight{1}%
    \setlength\tabcolsep{0pt}%
    \put(0,0){\includegraphics[width=\unitlength,page=1]{shapes_supp_qual.pdf}}%
    \put(0.04986878,0.86790459){\color[rgb]{0,0,0}\makebox(0,0)[lt]{\lineheight{1.25}\smash{\begin{tabular}[t]{l}Input\end{tabular}}}}%
    \put(0.17424381,0.86790483){\color[rgb]{0,0,0}\makebox(0,0)[lt]{\lineheight{1.25}\smash{\begin{tabular}[t]{l}CycleGAN\end{tabular}}}}%
    \put(0.34417879,0.86536331){\color[rgb]{0,0,0}\makebox(0,0)[lt]{\lineheight{1.25}\smash{\begin{tabular}[t]{l}UNIT\end{tabular}}}}%
    \put(0.47506231,0.86534997){\color[rgb]{0,0,0}\makebox(0,0)[lt]{\lineheight{1.25}\smash{\begin{tabular}[t]{l}AGGAN\end{tabular}}}}%
    \put(0.60021189,0.86536331){\color[rgb]{0,0,0}\makebox(0,0)[lt]{\lineheight{1.25}\smash{\begin{tabular}[t]{l}InstaGAN\end{tabular}}}}%
    \put(0.7638943,0.86536331){\color[rgb]{0,0,0}\makebox(0,0)[lt]{\lineheight{1.25}\smash{\begin{tabular}[t]{l}Ours\end{tabular}}}}%
    \put(0.86684294,0.86536331){\color[rgb]{0,0,0}\makebox(0,0)[lt]{\lineheight{1.25}\smash{\begin{tabular}[t]{l}Ground Truth\end{tabular}}}}%
    \put(0,0){\includegraphics[width=\unitlength,page=2]{shapes_supp_qual.pdf}}%
  \end{picture}%
\endgroup%

%% file: shapes_supp_qual_reverse.pdf_tex
\begingroup%
  \makeatletter%
  \providecommand\color[2][]{%
    \errmessage{(Inkscape) Color is used for the text in Inkscape, but the package 'color.sty' is not loaded}%
    \renewcommand\color[2][]{}%
  }%
  \providecommand\transparent[1]{%
    \errmessage{(Inkscape) Transparency is used (non-zero) for the text in Inkscape, but the package 'transparent.sty' is not loaded}%
    \renewcommand\transparent[1]{}%
  }%
  \providecommand\rotatebox[2]{#2}%
  \newcommand*\fsize{\dimexpr\f@size pt\relax}%
  \newcommand*\lineheight[1]{\fontsize{\fsize}{#1\fsize}\selectfont}%
  \ifx\svgwidth\undefined%
    \setlength{\unitlength}{878.38128662bp}%
    \ifx\svgscale\undefined%
      \relax%
    \else%
      \setlength{\unitlength}{\unitlength * \real{\svgscale}}%
    \fi%
  \else%
    \setlength{\unitlength}{\svgwidth}%
  \fi%
  \global\let\svgwidth\undefined%
  \global\let\svgscale\undefined%
  \makeatother%
  \begin{picture}(1,0.8832117)%
    \lineheight{1}%
    \setlength\tabcolsep{0pt}%
    \put(0,0){\includegraphics[width=\unitlength,page=1]{shapes_supp_qual_reverse.pdf}}%
    \put(0.04986878,0.86790459){\color[rgb]{0,0,0}\makebox(0,0)[lt]{\lineheight{1.25}\smash{\begin{tabular}[t]{l}Input\end{tabular}}}}%
    \put(0.17424381,0.86790483){\color[rgb]{0,0,0}\makebox(0,0)[lt]{\lineheight{1.25}\smash{\begin{tabular}[t]{l}CycleGAN\end{tabular}}}}%
    \put(0.34417879,0.86536331){\color[rgb]{0,0,0}\makebox(0,0)[lt]{\lineheight{1.25}\smash{\begin{tabular}[t]{l}UNIT\end{tabular}}}}%
    \put(0.47506231,0.86534997){\color[rgb]{0,0,0}\makebox(0,0)[lt]{\lineheight{1.25}\smash{\begin{tabular}[t]{l}AGGAN\end{tabular}}}}%
    \put(0.60021189,0.86536331){\color[rgb]{0,0,0}\makebox(0,0)[lt]{\lineheight{1.25}\smash{\begin{tabular}[t]{l}InstaGAN\end{tabular}}}}%
    \put(0.7638943,0.86536331){\color[rgb]{0,0,0}\makebox(0,0)[lt]{\lineheight{1.25}\smash{\begin{tabular}[t]{l}Ours\end{tabular}}}}%
    \put(0.86684294,0.86536331){\color[rgb]{0,0,0}\makebox(0,0)[lt]{\lineheight{1.25}\smash{\begin{tabular}[t]{l}Ground Truth\end{tabular}}}}%
    \put(0,0){\includegraphics[width=\unitlength,page=2]{shapes_supp_qual_reverse.pdf}}%
  \end{picture}%
\endgroup%